\ifdefined\pdfinfoomitdate\pdfinfoomitdate=1\fi
\ifdefined\pdftrailerid\pdftrailerid{}\fi
\ifdefined\pdfsuppressptexinfo\pdfsuppressptexinfo=15\fi
\documentclass[10pt,twocolumn,letterpaper]{article}
\usepackage[pagenumbers]{cvpr}

\usepackage{cuted} 

\usepackage{currfile} 

\usepackage{caption} 

\usepackage[utf8]{inputenc}
\usepackage[T1]{fontenc}
\usepackage{amsmath,amssymb,amsfonts}
\usepackage{url}
\usepackage{booktabs}
\usepackage{microtype}
\usepackage{xcolor}
\usepackage{colortbl}
\usepackage{graphicx}
\usepackage{pgfplots}
\pgfplotsset{compat=1.18}
\usepackage{subcaption}
\usepackage{multirow}
\usepackage{xspace}
\usepackage{array}
\usepackage{enumitem}
\usepackage{placeins}
\usepackage{capt-of}
\usepackage[most]{tcolorbox}
\usepackage{algorithm}
\usepackage{algpseudocode}

\tcbset{vmembox/.style={enhanced,colback=black!2,colframe=black!35,boxrule=0.45pt,
  sharp corners,fonttitle=\small\bfseries,fontupper=\small,
  top=1.5mm,bottom=1.5mm,left=2mm,right=2mm}}

\definecolor{cvprblue}{rgb}{0.21,0.49,0.74}
\usepackage[pagebackref,breaklinks,colorlinks,hypertexnames=false,allcolors=cvprblue]{hyperref}
\usepackage[capitalize,noabbrev]{cleveref}

\def\confName{CVPR}
\def\confYear{2027}

\graphicspath{{./}{figures/}{tables/}}
\newcommand{\bench}{VMem-Bench\xspace}

\providecommand{\dblender}{BlenderOpenMovies\xspace}
\providecommand{\dlsmdc}{LSMDC\xspace}

\definecolor{sysTF}{HTML}{2F5A63}
\definecolor{sysSU}{HTML}{6B8F5E}
\definecolor{sysSD}{HTML}{C4A35A}
\definecolor{sysSF}{HTML}{4E93B0}
\definecolor{sysLL}{HTML}{7A6499}
\definecolor{sysMF}{HTML}{6B3226}
\definecolor{sysIAM}{HTML}{C0684F}

\newcommand{\ci}[2]{\mbox{#1\ensuremath{_{\textcolor{black!55}{\scriptscriptstyle #2}}}}}

\newcommand{\VMemTeaser}{%
\vspace{-0.2in}%
\begin{center}
\includegraphics[width=\linewidth]{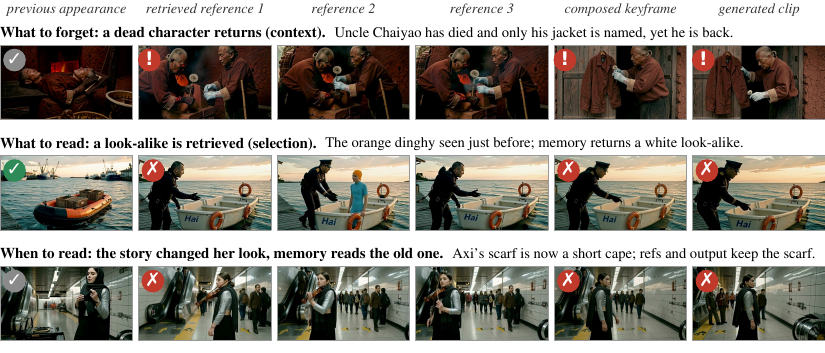}
\captionof{figure}{\textbf{A relevant past can be wrong for the present.}
Three qualitative failures from earlier Track-B testbed rollouts
(not the main quantitative rerun), showing previous appearances,
retrieved frames, composed keyframes, and generated clips.
\textbf{Top:} a deceased character reappears when a jacket is mentioned.
\textbf{Middle:} a white look-alike replaces the requested orange dinghy.
\textbf{Bottom:} retrieved frames preserve Axi's old scarf despite a
new instruction specifying a short cape. Symbols indicate
required-entity matches (\checkmark), misses ($\times$), and
forbidden appearances (!). These examples illustrate failure
paths; they are not fixed-history causal interventions.}
\label{fig:examples}
\end{center}
\vspace{-0.05in}}

\patchcmd{\maketitlesupplementary}
  {\vspace{1.0em}}
  {\vspace{1.0em}%
  \par\noindent
  \begin{minipage}[t]{\columnwidth}
    \vspace{0pt}\centering
    \captionsetup{hypcap=false}
    \includegraphics[width=\linewidth]{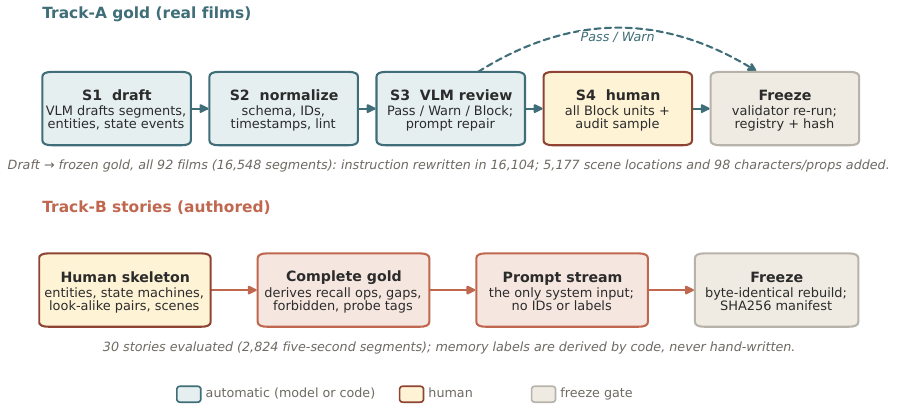}
\captionof{figure}{\textbf{How the two tracks are built and frozen.} Track-A gold starts from a VLM draft of each real
film; deterministic normalization and a VLM review route every blocked unit, plus a stratified sample of the
rest, to blocking human review before the full validator re-runs and the gold is frozen. Track-B stories are
authored: a human writes only the semantic skeleton, code derives every memory label and the prompt stream,
and a story is frozen only if both rebuild byte-identically from the skeleton.}
\label{fig:app-construction}

  \end{minipage}\hfill
  \begin{minipage}[t]{\columnwidth}
    \vspace{0pt}\centering
    \captionsetup{hypcap=false}
    \includegraphics[width=\linewidth]{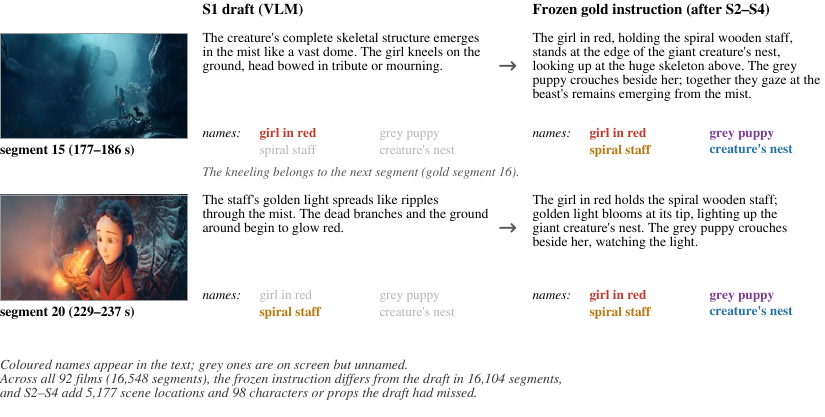}
\captionof{figure}{\textbf{From S1 draft to frozen gold.} In two \emph{Spring} segments, the VLM draft describes the
action but leaves entities that are on screen unnamed, and the first draft also borrows an action from the
next segment. The frozen instruction names every present entity by its roster name, so a memory can be asked
for each of them. The footer counts the same kind of change over the whole corpus.}
\label{fig:app-tracka-draft-gold}

  \end{minipage}
  \par\vspace{1em}%
}
  {}
  {\PackageError{vmem-paper}{Supplementary opening hook was not found}
    {Check the supplementary title macro in the bundled CVPR style.}}

\makeatletter
\g@addto@macro\@maketitle{\VMemTeaser}
\makeatother

\let\VMemAbsBegin\abstract
\def\VMemAbsEnd{\vspace*{0pt}}
\renewcommand{\abstract}[1]{%
  \begingroup
  \VMemAbsBegin #1\VMemAbsEnd
  \endgroup
}

\title{More Memory, Better Videos? A Controlled Study of Visual Memory in Long-Video Generation}
\author{%
  {\normalsize Yuzhuo~Chen$^{1,2,\dagger}$\quad
  Huafeng~Shi$^{2,\star,\ddagger}$\quad
  Xinyu~Wang$^{2,3}$\quad
  Yulong~Xu$^{2}$}\\[0pt]
  {\normalsize Yucheng~Wang$^{2,4}$\quad
  Haoqin~Hong$^{1,2,5}$\quad
  Guoxin~Zhang$^{2}$\quad
  Zehua~Ma$^{1,\star}$\quad
  Weiming~Zhang$^{1}$}\\[2pt]
  {\small $^{1}$University of Science and Technology of China\quad
  $^{2}$KlingAI Research\quad $^{3}$Tsinghua University}\\[0pt]
  {\small $^{4}$The Hong Kong University of Science and Technology\quad
  $^{5}$City University of Hong Kong}\\[2pt]
  {\small $^{\star}$Co-corresponding authors\quad $^{\ddagger}$Project leader}\\[0pt]
  {\small $^{\dagger}$This work was conducted during the author's internship at KlingAI}%
}

\makeatletter
\patchcmd{\@maketitle}{\vspace*{24pt}}{\vspace*{8pt}}{}{%
  \PackageError{arxiv-export}{Public title spacing anchor changed}{Update export_arxiv.py for the new style.}}
\makeatother
\begin{document}
\maketitle
\abstract{%
Long-video generators increasingly retrieve past visual observations to
prevent identity drift. The intuition is compelling: more access to history
should make future video more consistent. But an image can be faithful to
the past and wrong for the present. \textbf{Does visual memory always
improve the generated story?} We investigate this question through
\textbf{\bench}, a two-track diagnostic study. Track-A measures which
references are selected from 92 real films (16,548 segments), before
generation and without generator confounds. Track-B compares four retrieval
policies with no-memory and gold-informed last-seen controls on 30
causally generated stories (2,824 segments) using a shared generator.
Retrieval increases entity recall from 0.559 to 0.628--0.693 and improves
similarity to past appearances, but decreases precision from 0.977 to
0.915--0.938 and forbidden-entity avoidance from 0.968 to 0.845--0.898.
It can also reinforce obsolete appearances after the story changes them.
Conversely, high long-gap recall can arise without memory when the prompt
itself supplies the appearance; an appearance-stripping analysis exposes
this confound. These results distinguish \emph{historical fidelity} from
\emph{current-story correctness} and show why retrieval relevance or
final-video consistency alone does not establish memory utility.
\par\noindent{\normalfont\small\setlength{\baselineskip}{10pt}%
\textbullet\ Dataset: \href{https://huggingface.co/datasets/Suchenl/VMem-Bench}{huggingface.co/datasets/Suchenl/VMem-Bench}\\
\textbullet\ Code: \href{https://github.com/Suchenl/VMem-Bench}{github.com/Suchenl/VMem-Bench}\\
\textbullet\ Project Page: \href{https://suchenl.github.io/VMem-Bench/}{suchenl.github.io/VMem-Bench}
\par}
}
\section{Introduction}

\emph{More memory, better videos?} Causal long-video generation
produces a sequence of segments while carrying past observations
forward~\citep{yang2025longlive,yuan2026helios,meng2026causalcine}.
Historical retrieval can restore visual context when a character
returns~\citep{yu2025contextmemory,hu2026longliverag}.
Other systems compress frame or KV histories~\citep{zhang2025framepack,ji2025memflow},
or organize memory by entity~\citep{liu2026iamflow,dou2026slotmemory}.
Memory-conditioned storytelling likewise uses earlier shots to anchor
later appearances~\citep{zhang2025storymem,wang2026subjectanchor}.
But the story may have changed: the character wears a new outfit,
an object has been destroyed, or a deceased person is mentioned
without being present. A retrieved frame can improve \emph{fidelity
to history} while making the current segment \emph{less correct}
(Fig.~\ref{fig:examples}).

Beyond perceptual and cross-shot consistency
measured by existing evaluations
\citep{huang2023vbench,wu2025moviebench,he2026entitybench}:
\textbf{when does access to visual history actually help generation,
and when does it introduce interference?}
Three confounds complicate this question.
\emph{First}, the current prompt may already describe a returning entity.
\emph{Second}, historical matching may preserve obsolete appearances;
state tracking differs from identity anchoring~\citep{liu2026iamflow,zhou2026videomemory}.
\emph{Third}, relevant references may contain now-forbidden entities.
Retrieval evaluation separates relevance, downstream utility, and
distracting evidence
\citep{ru2024ragchecker}. For visual generation, current-state
correctness and entity exclusion must be measured alongside
historical resemblance.

We use \textbf{\bench} to separate \emph{what
memory retrieves} from \emph{what the generator does with it}.
At step $n$, a policy reads references $\widehat{\mathcal C}_n$
from historical memory $\mathcal M_n$ for instruction $g_n$;
the generator then produces $x_n$:
\begin{equation}
\mathcal M_n \xrightarrow{\;R(g_n)\;}
\widehat{\mathcal C}_n \xrightarrow{\;G(g_n)\;} x_n.
\label{eq:memory-transfer}
\end{equation}
\textbf{Track-A} replays 92 real films, evaluating reference
selection against frozen entity-level gold before the next
segment is revealed. \textbf{Track-B} runs 30 authored stories
causally, comparing four retrieval policies with no-memory
and gold-informed last-seen controls under one video-generation
pipeline. These matched-prompt comparisons evaluate \emph{complete policies}:
each policy writes its own history. They are not single-step interventions,
and the two tracks use different instances.

\paragraph{An unexpected answer.}
Within this controlled setting, retrieval improves overall
entity recall and appearance similarity to historical renderings.
Yet it also decreases precision and forbidden-entity avoidance;
on changed-state returns it can restore an outdated appearance.
Long-gap success partly reflects prompt information: no-memory
generation can achieve high recall when instructions describe the entity. Memory utility therefore requires checking \emph{what history adds},
\emph{which temporal state is valid}, and \emph{what unwanted content
it brings}.
These findings concern visual-history utility, not a memory-capacity scaling law.

\paragraph{Contributions.}
\begin{itemize}[leftmargin=1.25em,itemsep=1pt,topsep=2pt]
\item \textbf{A controlled characterization of memory's trade-offs.}
We show, across four retrieval policies and two controls, that
historical references increase recall and appearance continuity
while also increasing unwanted appearances and sometimes
conflicting with updated story states.
\item \textbf{Evidence of an evaluation confound.}
A targeted appearance-stripping analysis shows why successful
long-gap reappearance can reflect prompt information rather
than long-range memory. We separate historical appearance
similarity, present-state correctness, and entity exclusion.
\item \textbf{A reusable diagnostic study framework.}
\bench combines pre-generation evidence analysis (92 films,
16,548 segments, 2,803 entities) and post-generation,
closed-loop evaluation (30 stories, 2,824 segments, nine memory
stresses), with controlled retrieval comparisons, frozen
scoring, and explicit limitations on attribution.
\end{itemize}

\section{Related Work}

\paragraph{Visual memory for video generation.}
Visual histories support causal generation through compressed frame or KV
context~\citep{zhang2025framepack,ji2025memflow,yi2026worldkv}, historical
retrieval~\citep{yu2025contextmemory,hu2026longliverag,zhao2026omnimem},
and entity-indexed state~\citep{liu2026iamflow,dou2026slotmemory,zhou2026videomemory}.
OneStory learns adaptive historical-frame conditioning~\citep{an2025onestory};
StoryMem learns memory-conditioned multi-shot generation~\citep{zhang2025storymem}.
LiveEdit studies streaming editing with content-preservation
constraints~\citep{wang2026liveedit}; a broader survey organizes controllable
video conditioning mechanisms~\citep{ma2025controllablesurvey}.
Holding the generator fixed across retrieval policies, we test whether
reference relevance and historical resemblance predict current-story
correctness.

\paragraph{Long-video and memory benchmarks.}
VBench measures perceptual and semantic quality~\citep{huang2023vbench}.
MovieBench, MSVBench, and EntityBench evaluate movie-level or cross-shot
consistency~\citep{wu2025moviebench,shi2026msvbench,he2026entitybench};
MBench and NarraStream-Bench test memory or narrative
consistency~\citep{zhang2026mbench,liu2026iamflow}.
SubjectAnchor combines subject-aware memory with a trained video model and
its own storytelling benchmark, including shot-level inclusion/exclusion
control~\citep{wang2026subjectanchor}.
Our two tracks measure selected evidence before generation and story
correctness afterward, complementing these evaluations and SubjectAnchor's
inclusion/exclusion tests. This does not establish superiority over its
trained architecture (Supplementary Table~\ref{tab:nearby-memory-work};
App.~\ref{app:system-adaptation}).

\paragraph{Retrieval relevance and downstream utility.}
LongMemEval and MemoryAgentBench connect memory operations to task
success~\citep{wu2025longmemeval,hu2026memoryagentbench}.
RAGChecker already separates retrieval quality from generation quality and
identifies distracting evidence~\citep{ru2024ragchecker}.
Our question concerns temporal validity in visual generation: historical
appearance continuity can coexist with an incorrect current state or an
unwanted entity. We jointly test selected evidence, updated states,
forbidden appearances, and prompt-information sensitivity. No-memory
controls check what history adds beyond the current instruction, while
complete-policy comparisons measure the outcome of each evolving memory
history, with explicit limits on causal attribution.

\section{Study Design: Evidence and Outcomes}
\label{sec:bench}

To study the utility of history, we observe two points in
Eq.~\ref{eq:memory-transfer}: the references a policy selects before
generation (Track-A) and the video produced afterward (Track-B).
\bench is the experimental instrument for this investigation;
Fig.~\ref{fig:bench-run} shows the two protocols.
For example, Track-A asks whether selected references cover a puppy requested
in a later segment of the real film \emph{Spring} (Fig.~\ref{fig:app-tracka-refs}).
Track-B instead tests memory use in independently authored, generated stories,
with no-memory and last-seen controls. The two tracks measure complementary
boundaries, not matched instances.

\subsection{Causal Evaluation Contract}
\label{sec:bench-protocol}
Here causal means no access to future segments. Both tracks process a story chronologically; they differ in what is scored.

\begin{algorithm}[!ht]
\caption{Causal evaluation contract}
\label{alg:eval-contract}
\small
\begin{algorithmic}[1]
\Require Track $T\in\{\mathrm{A},\mathrm{B}\}$; instruction stream $g_{1:N}$; policy $\pi$; scoring gold $G$
\Ensure Per-segment references and scores; $G$ is hidden from ordinary policies; gold-ID Last-seen is a privileged Track-B diagnostic
\State Initialize memory $\mathcal{M}_1$ from no prior segments
\For{$n=1,\ldots,N$}
  \If{$T=\mathrm{B}$ and $\pi$ is gold-ID Last-seen}
    \State Select up to $K$ prior frames using gold required-entity IDs and this rollout's earlier clips
  \Else
    \State Give ordinary $\pi$ only $g_n$ and its earlier-segment memory $\mathcal{M}_n$
    \State Select reference context $\widehat{\mathcal{C}}_n$ with budget $B_n=|\widehat{\mathcal{C}}_n|$
  \EndIf
  \If{$T=\mathrm{A}$ (real-film retrieval)}
    \State Reveal the next real segment only after selection; score $\widehat{\mathcal{C}}_n$ against its entities and states
    \State Observe the current real segment and update memory; no generator runs
  \Else \Comment{$T=\mathrm{B}$ (closed-loop generation)}
    \State Generate $x_n$ from $g_n$ and $\widehat{\mathcal{C}}_n$; score entities and states against blinded gold
    \State Observe the system's own $x_n$ and update $\mathcal{M}_{n+1}$
  \EndIf
\EndFor
\end{algorithmic}
\end{algorithm}

Track-A measures the evidence handed to a generator, without running one; its instructions name the entities on screen.
Track-B measures what survives generation, with a shared pipeline across conditions (Sec.~\ref{sec:exp-setting}).
Because each condition updates memory from its own output, their histories diverge during the closed-loop rollout.
Scoring gold describes entities and states; it is hidden from ordinary policies and used only for scoring, except for the explicitly privileged gold-ID Last-seen diagnostic (App.~\ref{app:last-seen}). It supplies no reference pixels or descriptions.

\subsection{Data, Gold, and Memory Stresses}
\label{sec:bench-construction}
Track-A is built from \dblender{} and \dlsmdc{} (Table~\ref{tab:bench-stats}): 92 films, 16,548 segments,
44.91 hours, and 2,803 typed characters, props, and locations, with reappearance,
multi-instance, scene-return, and state-change slices.

\begin{figure*}[!t]
\centering
\resizebox{0.94\textwidth}{!}{%
  \includegraphics[height=4.55cm,keepaspectratio]{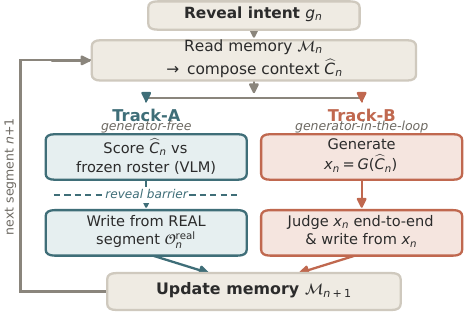}%
  \hspace{0.6em}%
  \includegraphics[height=4.55cm,keepaspectratio]{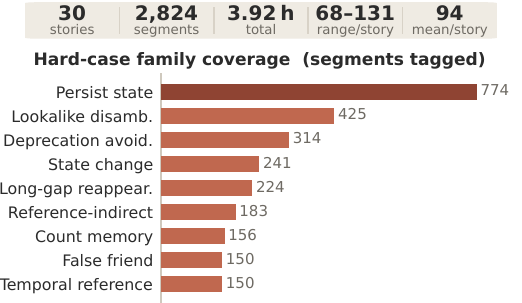}%
}%
\begin{subfigure}{\textwidth}
\phantomcaption\label{fig:bench-run}%
\phantomcaption\label{fig:trackb-stats}%
\end{subfigure}

\captionsetup{singlelinecheck=false,justification=justified,font=small}
\caption{\textbf{(a)~Causal evaluation.} Both tracks read references for the current instruction. Track-A scores references before revealing the real next segment, then updates from that segment. Track-B generates and scores a clip, then updates from its own output. \textbf{(b)~Track-B cohort and stress coverage.} A segment can carry multiple stress tags.}

\end{figure*}

Track-B contains 30 human-authored stories expanded into 2,824 five-second segments (3.92 hours), with
reappearances up to 77 segments apart. The instruction gives an entity's look only when it first appears or
changes state; later prompts usually name the entity without repeating its full appearance. Many props,
however, carry their look in the name (``the two brass fog bells''), and most returns after 30 or more
segments are either such props or characters that come back changed, with the new look spelled out. Each segment is tagged with the memory stresses it exercises
(Fig.~\ref{fig:trackb-stats}): six ask the system to \emph{remember} something and three to \emph{keep
something out} (Table~\ref{tab:trackb-families}).

\begin{table*}[!t]
\centering
\caption{Track-B memory-stress families. Each rate is a micro-average over that family's opportunities.}
\label{tab:trackb-families}
\footnotesize
\setlength{\tabcolsep}{3.5pt}
\renewcommand{\arraystretch}{1.02}
\begin{tabular}{@{}l l p{0.55\textwidth}@{}}
\toprule
Family & Metric & Satisfied when \\
\midrule
Long-gap reappearance & recall & A character or prop returns after at least 30 segments. \\
Look-alike & accuracy & The identity matches the intended member of a confusable pair. \\
State change & accuracy & A transformed entity appears in its new state. \\
Count / multi-instance & accuracy & The rendered count matches the authored count. \\
Avoidance & avoidance-OK & A removed, destroyed, or deceased entity is absent. \\
False friend & accuracy & A new look-alike is not collapsed into an existing identity. \\
State persistence & accuracy & A later segment keeps an earlier transformed state. \\
Indirect reference & avoidance-OK & A deprecated entity stays absent even if named in the prose. \\
Temporal reference & recall & A name-free time reference recovers the intended entity. \\
\bottomrule
\end{tabular}
\end{table*}

Track-A gold comes from a staged pipeline: VLM proposals are normalized by deterministic contracts and
stronger-model review triggers targeted repair. Humans fix the canonical entity roster, and every blocked
unit plus a stratified audit sample of the rest passes blocking human review before a deterministic
freeze (App.~\ref{app:bench-making}).

\begin{table*}[!t]
\centering
\caption{\textbf{\bench{} Track-A corpus statistics} (per dataset). Entities =
characters $+$ props $+$ locations; hard-case columns count segments per bucket.
\dblender is stylized animation, \dlsmdc live-action film clips; the two are analyzed
separately throughout.}
\label{tab:bench-stats}
\small
\setlength{\tabcolsep}{3.5pt}
\resizebox{0.90\textwidth}{!}{%
\begin{tabular}{@{}l rrr rrr r rrrr@{}}
\toprule
& \multicolumn{7}{c}{\textbf{Corpus}} & \multicolumn{4}{c}{\textbf{Hard-case segments}} \\
\cmidrule(lr){2-8}\cmidrule(l){9-12}
Dataset & Movies & Segments & Entities & Char. & Prop. & Loc. & Dur.\,(h)
        & Re-app. & Multi-inst. & Scene-ret. & State-chg. \\
\midrule
\dblender & $23$ & $1{,}364$ & $369$ & $106$ & $149$ & $114$ & $3.84$ & $415$ & $674$ & $106$ & $86$ \\
\dlsmdc   & $69$ & $15{,}184$ & $2{,}434$ & $881$ & $690$ & $863$ & $41.07$ & $5{,}425$ & $8{,}596$ & $1{,}582$ & $179$ \\
\midrule
Total     & $92$ & $16{,}548$ & $2{,}803$ & $987$ & $839$ & $977$ & $44.91$ & $5{,}840$ & $9{,}270$ & $1{,}688$ & $265$ \\
\bottomrule
\end{tabular}%
}
\end{table*}

\subsection{Metrics}
\label{sec:bench-metrics}
\paragraph{Track-A.}
A pinned Qwen3-VL-32B judge, blind to system identity, checks each reference against the frozen roster
and the real next segment. At segment $n$, let $C_n$ be the entities that must be remembered (on screen,
excluding first appearances) and $k_n$ the number of references. Precision is the fraction of the $k_n$
references that show a correct entity; Recall is the fraction of $C_n$ covered by at least one of them.
Both, and their F1, are computed per segment and averaged over a film's segments, then over films (so the
reported F1 is not the harmonic mean of the reported averages). Repetition is measured as $R_V$, the share
of references that repeat an already covered view of the same entity, and $R_S$, their mean DINOv3
similarity; $\mathrm{SelEff}=\mathrm{Prec}\,(1-R_V)$.

\paragraph{Track-B.}
The same judge sees each generated segment with a blinded list of entities that should appear, should not
appear, look alike, or are decoys, each with its gold description, and marks which are present. An entity
counts as present when the clip shows something that matches its description; the judge does not compare it
with the entity's earlier renderings, so a fresh, description-matching stranger scores as a hit. Track-B
therefore also scores \emph{identity}: whether a returning entity looks the same as its previous appearance
in the same rollout, measured as the DINOv3 similarity of the localized entity over returns detected in both
clips, with ArcFace for faces, MegaLoc for locations, and a pairwise VLM judge as a check; thresholds are
calibrated on control pairs before any condition is scored (App.~\ref{app:identity}). Counts are pooled over
all segments.
Recall ($R$) is the fraction of required character and prop appearances
that are present; Precision ($P$) is the fraction of present characters and
props that were required; F1 is their harmonic mean; Avoid-OK
($A$) is the fraction of forbidden appearances that stay absent.
Rather than collapse memory usefulness into a single consistency
score, we examine the vector of \emph{policy-level} differences
to no memory:
\begin{equation}
\Delta_{\pi}=(R_\pi-R_{\varnothing},\
P_\pi-P_{\varnothing},\
A_\pi-A_{\varnothing}).
\label{eq:policy-effects}
\end{equation}
Here $\pi$ denotes a memory policy and $\varnothing$ denotes
the no-memory condition. Each stress family has its own rate.
Paired story-clustered bootstrap 95\% CIs describe variation
across stories conditional on one rollout per condition/story
(App.~\ref{app:bench-scoring}). Because each condition writes
its own generated history, $\Delta_\pi$ compares whole policies,
not isolated reference choices within an identical history.

\begin{figure*}[t]
\centering
\includegraphics[width=\textwidth]{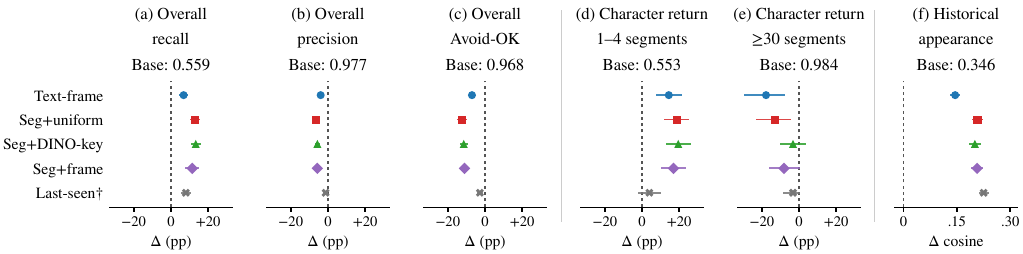}
\caption{\textbf{Overall memory trade-offs and diagnostic slices.} Points: policy minus no memory (zero line); whiskers: paired story-bootstrap 95\% CIs conditional on one rollout/story. Titles give absolute no-memory baselines.
\textbf{(a,b)} Pooled character/prop Recall and Precision; \textbf{(c)} forbidden-entity Avoid-OK. \textbf{(d,e)} Character-return recall; 61/62 long-gap returns change state with prompt-described looks. All five use the same percentage-point scale.
\textbf{(f)} Historical DINOv3 cosine on a separate scale, on policy-dependent subsets detected in both appearances; higher need not mean more correct.
\textsuperscript{†}Gold-ID Last-seen uses a variable budget; retrieval uses $K{=}3$.}
\label{fig:references-effect}
\end{figure*}

\begin{table*}[!t]
\definecolor{memup}{HTML}{E0ECF4}\definecolor{memdown}{HTML}{F6DDD5}
\centering
\caption{\textbf{Track-B: generated-video outcomes by stress} (30 stories/system; higher is better). Cells are absolute rates. In shared-generator policy rows, blue/+ and red/$-$ mark paired story-bootstrap 95\% CIs above/below the first-row no-memory baseline; unmarked cells have no supported direction. Long-gap pools characters and props; Table~\ref{tab:main-kind-gap} separates them. Overall is pooled character/prop recall. The separate released-system block uses different generators and is not a memory-only comparison. Stress-level judge sensitivity: Sec.~\ref{sec:exp-sensitivity}.}
\label{tab:trackb-results}
\small
\setlength{\tabcolsep}{3pt}
\renewcommand{\arraystretch}{1.06}
\begin{tabular}{@{}l cccccccccc@{}}
\toprule
& \multicolumn{5}{c}{\textbf{Presence, identity, and count}} & \multicolumn{2}{c}{\textbf{State correctness}} & \multicolumn{2}{c}{\textbf{Entity exclusion}} & \textbf{All} \\
\cmidrule(lr){2-6}\cmidrule(lr){7-8}\cmidrule(lr){9-10}\cmidrule(lr){11-11}
System & Long-gap & Look-alike & False-fr. & Count & Temporal & State-chg & Persist & Avoid & Indirect & Overall$\uparrow$ \\
 & {rec.} & {acc.} & {acc.} & {acc.} & {rec.} & {acc.} & {acc.} & {avoid.} & {avoid.} & {rec.} \\
\midrule
\multicolumn{11}{@{}l}{\itshape Shared generator: no-memory baseline, four retrieval policies, Last-seen control}\\
No memory ($K{=}0$) & 0.890 & 0.952 & 0.980 & 0.779 & 0.867 & 0.991 & 0.689 & 0.964 & 0.852 & 0.559 \\
Text-frame ret. & 0.819 & 0.964 & 0.960 & 0.803 & 0.827 & 0.995 & \cellcolor{memup}0.802\textsuperscript{+} & \cellcolor{memdown}0.889\textsuperscript{$-$} & \cellcolor{memdown}0.694\textsuperscript{$-$} & \cellcolor{memup}0.628\textsuperscript{+} \\
Seg+uniform ret. & 0.871 & 0.959 & \cellcolor{memdown}0.900\textsuperscript{$-$} & 0.794 & 0.860 & 0.991 & \cellcolor{memup}0.783\textsuperscript{+} & \cellcolor{memdown}0.829\textsuperscript{$-$} & \cellcolor{memdown}0.705\textsuperscript{$-$} & \cellcolor{memup}0.689\textsuperscript{+} \\
Seg+DINO-key ret. & 0.905 & 0.980 & \cellcolor{memdown}0.893\textsuperscript{$-$} & 0.819 & 0.893 & 0.991 & \cellcolor{memup}0.772\textsuperscript{+} & \cellcolor{memdown}0.839\textsuperscript{$-$} & \cellcolor{memdown}0.694\textsuperscript{$-$} & \cellcolor{memup}0.693\textsuperscript{+} \\
Seg+frame ret. & 0.871 & \cellcolor{memup}0.983\textsuperscript{+} & \cellcolor{memdown}0.920\textsuperscript{$-$} & 0.821 & 0.887 & 0.986 & \cellcolor{memup}0.797\textsuperscript{+} & \cellcolor{memdown}0.843\textsuperscript{$-$} & \cellcolor{memdown}0.628\textsuperscript{$-$} & \cellcolor{memup}0.674\textsuperscript{+} \\
Last-seen refs & 0.881 & 0.966 & 0.967 & 0.783 & \cellcolor{memdown}0.740\textsuperscript{$-$} & 0.996 & \cellcolor{memup}0.778\textsuperscript{+} & \cellcolor{memdown}0.939\textsuperscript{$-$} & 0.809 & \cellcolor{memup}0.639\textsuperscript{+} \\
\cmidrule(lr){1-11}
\rowcolor{black!5}\multicolumn{11}{@{}l}{\itshape Separate setting: released systems with their own generators}\\
MemFlow & 0.238 & 0.957 & 0.440 & 0.615 & 0.300 & 0.991 & 0.758 & 0.921 & 0.754 & 0.360 \\
LongLive-RAG & 0.386 & 0.981 & 0.740 & 0.718 & 0.327 & 0.994 & 0.712 & 0.983 & 0.929 & 0.284 \\
IAMFlow & 0.186 & 0.949 & 0.393 & 0.659 & 0.253 & 1.000 & 0.751 & 0.916 & 0.781 & 0.356 \\
\bottomrule
\end{tabular}
\end{table*}

\section{Experimental Findings}
\label{sec:experiments}

\subsection{Controlled Comparison}
\label{sec:exp-setting}
We evaluate four retrieval policies. \emph{Text-frame} ranks individual
historical frames by text similarity; \emph{Seg+uniform} retrieves
text-matched segments and samples frames uniformly; \emph{Seg+DINO-key}
chooses visually diverse DINOv3 keyframes from retrieved segments; and
\emph{Seg+frame} re-ranks frames within the top text-retrieved segments.
Track-A scores their selected references at a cap of $B{=}16$.
Track-B uses $K{=}3$ references from each condition's \emph{own
generated history}. One shared generator combines FLUX.2 keyframe
composition with Wan2.2-TI2V-5B five-second video generation.
We compare against
\emph{no memory} (no historical references) and \emph{last-seen}
(the most recent generated appearance of each required entity,
selected using gold entity IDs, but no gold pixels or descriptions).
Last-seen is a privileged diagnostic control, \emph{not} a deployable
baseline or a bound on achievable performance. Results use one
rollout per story and condition; paired story-clustered 95\% CIs
characterize variation across these 30 stories, conditional on
those rollouts (App.~\ref{app:bench-scoring}). Released systems form a separate native-generator block in
Table~\ref{tab:trackb-results}; generator differences preclude
memory-only comparisons (App.~\ref{app:system-adaptation}).

The evaluated cohort has 152 character, 300 prop, and 120 location
records; all 30 stories contain all nine stress families
(Supplementary Table~\ref{tab:story-coverage}). These authored probes do not represent
the distribution of movie genres.

\subsection{Finding I: Recall and Continuity Improve}
\label{sec:exp-distance}

On the shared generator, retrieval increases pooled character/prop
recall from $0.559$ without memory to $0.628$--$0.693$
(Table~\ref{tab:trackb-results}). For short-gap character returns (1--4 segments),
whose instructions typically identify the character by name,
no-memory recall is $0.553$, versus $0.698$--$0.749$ with
retrieval (Fig.~\ref{fig:references-effect}d). Table~\ref{tab:main-kind-gap}a separates entity types (Table~\ref{tab:trackb-results}
pools characters and props in its long-gap column): all four retrieval
policies reliably help short-gap characters, while none establishes
a long-gap gain for characters or props. None establishes a gain for
short-gap props either: pooling entity types can hide which returns
benefit from memory. These groups differ in prompt
content and state changes; they do not isolate an absence-gap effect.

\paragraph{A prompt-intervention check.}
Long absence alone does not demonstrate a need for memory: no-memory
recall rises from $0.58$ at gaps of 1--4 segments to $0.89$ at 30 or
more, where prompts more often specify an object's appearance or a
character's changed look. Table~\ref{tab:nolook}b tests the latter
confound on 62 long-gap character returns. Removing appearance
descriptions lowers no-memory presence from $0.984$ to $0.565$.
Seg+DINO-key then differs from no memory by only $+0.016$, with a
95\% CI spanning zero. This establishes prompt-information sensitivity, not a reliable
retrieval gain on the stripped subset: assess prompt information
before attributing long-gap success to memory.
\begin{table}[t]
\centering
\caption{\textbf{Return recall and prompt information} (30 stories). No memory: absolute recall; DINO-key $\Delta R$: difference from it. Gray subscripts: full paired story-bootstrap 95\% CI width; bold: CI excludes zero. $+/-/?$: counts of four retrieval policies with CIs above/below/including zero. Gaps: segments; prompts describe changed looks in 61/62 long-gap character returns. Only (b) changes prompts; see Supplementary Table~\ref{tab:kind-gap}.}
\label{tab:main-kind-gap}
\label{tab:nolook}
\footnotesize
\setlength{\tabcolsep}{3pt}
\begin{tabular}{@{}llrrcc@{}}
\toprule
\multicolumn{6}{@{}l}{\textit{(a) Original prompts: observational groups}}\\
Entity & Gap & $n$ & \shortstack{No\\memory} & \shortstack{DINO-key\\$\Delta R$} & \shortstack{Policies\\$+ / - / ?$} \\
\midrule
Character & 1--4 & 430 & 0.553 & \ci{\textbf{+0.195}}{0.120} & 4/0/0 \\
 & $\geq$30 & 62 & 0.984 & \ci{-0.032}{0.130} & 0/2/2 \\
\midrule
Prop & 1--4 & 271 & 0.635 & \ci{+0.041}{0.110} & 0/0/4 \\
 & $\geq$30 & 148 & 0.851 & \ci{+0.034}{0.190} & 0/0/4 \\
\bottomrule
\end{tabular}
\par\smallskip
\begin{tabular}{@{}lcccc@{}}
\toprule
\multicolumn{5}{@{}l}{\textit{(b) Appearance intervention: same 62 character returns}}\\
Prompt & \shortstack{No\\memory} & \shortstack{DINO-key\\$R$} & \shortstack{Last-seen\\$R$} & \shortstack{DINO-key\\$\Delta R$} \\
\midrule
Described & 0.984 & 0.952 & 0.952 & \ci{-0.032}{0.130} \\
Stripped & 0.565 & 0.581 & 0.565 & \ci{+0.016}{0.307} \\
\bottomrule
\end{tabular}
\end{table}

\paragraph{Continuity beyond presence.}
Memory also changes how a returning entity looks. Conditional on returning entities
localized in both appearances, the mean DINOv3 crop similarity to their
previous appearance is $0.35$ without memory and $0.49$--$0.57$
with retrieved references (Table~\ref{tab:main-identity}); similar gains
occur at longer gaps (App.~\ref{app:identity}). This is \emph{appearance continuity},
not proof that the appearance is correct for the entity's current
story state. Missed or unlocalized returns are excluded, and the
detected subset can differ across policies. A pairwise VLM judge also
reports continuity gains, but calls 78\% of different-entity negative-control
pairs the same individual when both entities are visible
(Supplementary Table~\ref{tab:identity-calibration}). It is therefore an
optimistic check (App.~\ref{app:identity}).

\begin{table}[t]
\centering
\caption{\textbf{Historical appearance continuity.} DINOv3 crop cosine: no-memory means and policy deltas ($\Delta$). Bold: paired story-bootstrap 95\% CI excludes zero. $N$ counts eligible events; detected pairs vary by policy. Changed: characters returning after $\geq$30 segments; similarity can reward obsolete looks.}
\label{tab:main-identity}
\footnotesize
\setlength{\tabcolsep}{2.5pt}
\begin{tabular}{@{}lrrrr@{}}
\toprule
Policy & \shortstack{All returns\\$N=1130$} & \shortstack{Characters\\$N=486$} & \shortstack{Props\\$N=644$} & \shortstack{Changed\\$N=61$} \\
\midrule
No memory & 0.35 & 0.27 & 0.41 & 0.19 \\
\midrule
Text-frame $\Delta$ & \textbf{+0.15} & \textbf{+0.13} & \textbf{+0.16} & \textbf{+0.18} \\
Seg+uniform $\Delta$ & \textbf{+0.21} & \textbf{+0.20} & \textbf{+0.21} & \textbf{+0.20} \\
Seg+DINO-key $\Delta$ & \textbf{+0.20} & \textbf{+0.19} & \textbf{+0.21} & \textbf{+0.20} \\
Seg+frame $\Delta$ & \textbf{+0.21} & \textbf{+0.18} & \textbf{+0.23} & \textbf{+0.15} \\
Last-seen $\Delta$ & \textbf{+0.23} & \textbf{+0.23} & \textbf{+0.22} & \textbf{+0.23} \\
\bottomrule
\end{tabular}
\end{table}

\subsection{Finding II: Past Fidelity Can Harm the Present}
\label{sec:exp-trackb}

\begin{figure}[t]
\centering
\includegraphics[width=\linewidth]{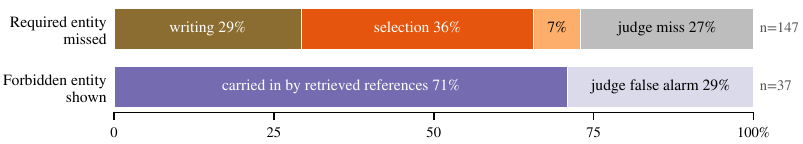}
\caption{\textbf{Track-B: where a failure arises}, from a stage-by-stage single-author audit of the four
retrieval memories in an earlier rollout (pooled, inclusion-weighted
conditional shares, not population failure rates). A missed required entity was already wrong in its previous
appearance (writing), present there but not retrieved (selection), retrieved but lost in the keyframe or clip
(synthesis), or rendered and missed by the judge. Forbidden-render units are reference-associated or judge false alarms;
association does not identify a causal reference effect. Per-stratum shares are in Fig.~\ref{fig:memory-effect}b.}
\label{fig:attribution}
\end{figure}

The benefit comes with a measurable cost. Pooled Precision drops
from $0.977$ without memory to $0.915$--$0.938$ with retrieval;
pooled Avoid-OK falls from $0.968$ to $0.845$--$0.898$
(Fig.~\ref{fig:references-effect}a--c). These changes have
paired story-bootstrap 95\% CIs excluding zero under the pinned judge.
For example, Seg+DINO-key has a $+0.133$ Recall gain
(95\% CI $+0.111$ to $+0.155$), but a $-0.057$ Precision
change ($-0.066$ to $-0.048$) and a $-0.114$ Avoid-OK change
($-0.132$ to $-0.098$). Report the signed effects in Eq.~\ref{eq:policy-effects} together:
recall gains can conceal unwanted appearances.
In cases where a destroyed or deceased entity is forbidden, the
deprecation-family Avoid-OK drops from $0.964$ to
$0.829$--$0.889$ (the deprecation-avoidance column in
Table~\ref{tab:trackb-results}). The generator may render an entity
carried in a relevant-looking historical frame, even when that
entity should no longer be present. A single-author internal audit of 224 selected units from an earlier
rollout traces errors to stored history, retrieval, synthesis, and judging
(Fig.~\ref{fig:attribution}; App.~\ref{app:human-attribution}).

More subtly, a reference can preserve the \emph{wrong} appearance.
Among 62 long-gap character returns, 61 change state with
an updated appearance described in the instruction
(App.~\ref{app:trackb-more}). Across all 62 returns, no-memory
presence according to the description reaches $0.984$; all four retrieval policies decrease it by
$0.03$--$0.18$, with two paired CIs below zero
(App.~\ref{app:trackb-more}). Their generated appearances move
closer to the \emph{previous} look, even though it is obsolete: on the
61 changed returns, DINOv3 similarity rises from $0.19$ by
$+0.15$--$+0.20$ (Table~\ref{tab:main-identity}). The 62-return
presence analysis includes one unchanged return; the two measures
also have different visibility conditions. Evaluate current-state correctness alongside historical resemblance;
matching the past does not establish correctness now.

The gold-informed last-seen control reduces, but does not eliminate,
reference interference: its pooled Recall/Precision/Avoid-OK are
$0.639/0.964/0.941$. Even an entity-matched historical frame can
contain a retired bystander or an outdated state. Its performance
must also be read alongside its smaller, variable reference count,
rather than interpreted as a budget-matched oracle. Table~\ref{tab:main-last-seen-budget}
separates its reference-count groups; their different segments preclude
causal count comparisons
(App.~\ref{app:last-seen}, App.~\ref{app:robustness}).
\begin{table}[t]
\centering
\caption{\textbf{Last-seen is not budget matched.} Deltas from no memory by references used. Gray subscripts: full paired story-bootstrap 95\% CI widths; bold: CI excludes zero. Different segments per group preclude causal count comparisons.}
\label{tab:main-last-seen-budget}
\footnotesize
\setlength{\tabcolsep}{3pt}
\begin{tabular}{@{}rrccc@{}}
\toprule
Refs & Segments & $\Delta R$ & $\Delta P$ & $\Delta A$ \\
\midrule
1 & 1,803 & \ci{\textbf{+0.087}}{0.044} & \ci{-0.002}{0.020} & \ci{-0.012}{0.028} \\
2 & 842 & \ci{\textbf{+0.078}}{0.063} & \ci{\textbf{-0.027}}{0.020} & \ci{\textbf{-0.047}}{0.032} \\
3 & 149 & \ci{+0.038}{0.092} & \ci{-0.031}{0.067} & \ci{-0.052}{0.104} \\
\bottomrule
\end{tabular}
\end{table}

Cross-judge support varies by policy; Sec.~\ref{sec:exp-sensitivity}
summarizes what replicates and the scope of the H3 backend check.

\subsection{Finding III: Relevance Is Not Utility}
\label{sec:exp-tracka}
\begin{table*}[!t]
\centering
\caption{\textbf{Track-A reference evidence at each memory's own budget} (Refs per segment). Precision/Recall/F1: correctness and entity coverage; $R_V/R_S$: repetition (App.~\ref{app:tracka-more}). SelEff $=\mathrm{Prec}(1-R_V)$. Underline: highest score/lowest repetition. Refs is not ranked.}
\label{tab:main-results}
\small
\setlength{\tabcolsep}{2.7pt}
\renewcommand{\arraystretch}{1.04}
\begin{tabular}{@{}l *{7}{c} @ {\hspace{7pt}} *{7}{c}@{}}
\toprule
\multirow{3}{*}{System}
& \multicolumn{7}{c}{\textbf{BlenderOpenMovies}} & \multicolumn{7}{c}{\textbf{LSMDC}} \\
\cmidrule(lr){2-8}\cmidrule(l){9-15}
& \multicolumn{3}{c}{Correctness / coverage} & \multicolumn{3}{c}{Efficiency / redundancy} & Budget
& \multicolumn{3}{c}{Correctness / coverage} & \multicolumn{3}{c}{Efficiency / redundancy} & Budget \\
& Prec$\uparrow$ & Rec$\uparrow$ & F1$\uparrow$ & SelEff$\uparrow$ & $R_V\downarrow$ & $R_S\downarrow$ & Refs
& Prec$\uparrow$ & Rec$\uparrow$ & F1$\uparrow$ & SelEff$\uparrow$ & $R_V\downarrow$ & $R_S\downarrow$ & Refs \\
\midrule
\multicolumn{15}{@{}l}{\itshape Retrieval memories (testbed)}\\
Text-frame ret.     & \underline{0.692} & 0.887 & \underline{0.817} & 0.323 & 0.492 & 0.634 & 15.47 & \underline{0.630} & 0.802 & \underline{0.670} & 0.330 & 0.404 & 0.543 & 15.94 \\
Seg+uniform ret.    & 0.662 & 0.883 & 0.796 & 0.297 & 0.515 & 0.650 & 15.47 & 0.566 & 0.769 & 0.622 & 0.264 & 0.454 & 0.605 & 15.92 \\
Seg+DINO-key ret.   & 0.579 & \underline{0.923} & 0.752 & \underline{0.431} & \underline{0.225} & \underline{0.488} & 13.80 & 0.484 & \underline{0.849} & 0.591 & \underline{0.393} & 0.143 & \underline{0.393} & 14.82 \\
Seg+frame ret.      & 0.667 & 0.869 & 0.792 & 0.294 & 0.525 & 0.637 & 15.47 & 0.580 & 0.772 & 0.630 & 0.285 & 0.433 & 0.573 & 15.93 \\
\cmidrule(lr){1-15}
\multicolumn{15}{@{}l}{\itshape Released memories}\\
LongLive-RAG        & 0.619 & 0.648 & 0.647 & 0.325 & 0.433 & 0.773 & 5.98 & 0.447 & 0.506 & 0.441 & 0.267 & 0.283 & 0.665 & 6.00 \\
MemFlow             & 0.321 & 0.597 & 0.422 & 0.194 & 0.294 & 0.625 & 15.94 & 0.240 & 0.530 & 0.302 & 0.185 & \underline{0.133} & 0.424 & 16.00 \\
\bottomrule
\end{tabular}
\end{table*}

\begin{table}[t]
\centering
\caption{\textbf{Clip quality versus entity recall.} Same 2,824 clips per condition; means with story-bootstrap 95\% CIs; gray subscripts show full widths (upper minus lower). All metrics are higher-is-better. Intervals condition on one rollout per story. Complete conditions and six quality metrics: Supplementary Table~\ref{tab:quality-control}.}
\label{tab:main-quality}
\footnotesize
\setlength{\tabcolsep}{3pt}
\begin{tabular}{@{}lcc@{}}
\toprule
Metric & No memory & Seg+DINO-key \\
\midrule
Imaging & \ci{0.645}{0.015} & \ci{0.652}{0.023} \\
Aesthetic & \ci{0.573}{0.013} & \ci{0.595}{0.024} \\
Subject consistency & \ci{0.935}{0.004} & \ci{0.929}{0.004} \\
Background consistency & \ci{0.941}{0.002} & \ci{0.931}{0.003} \\
Per-clip entity recall & \ci{0.554}{0.068} & \ci{0.740}{0.045} \\
\bottomrule
\end{tabular}
\end{table}

\begin{table}[t]
\centering
\caption{\textbf{Judge and backend sensitivity of Seg+DINO-key.} Deltas from no memory; gray subscripts give full paired story-bootstrap 95\% CI widths, conditional on one rollout per story. Cross-judge support counts policies with same-sign CIs excluding zero. Backend rows retain FLUX.2 keyframes. Limits: Sec.~\ref{sec:exp-sensitivity}.}
\label{tab:main-sensitivity}
\footnotesize
\setlength{\tabcolsep}{3pt}
\begin{tabular}{@{}lccc@{}}
\toprule
Judge / backend & $\Delta R$ & $\Delta P$ & $\Delta A$ \\
\midrule
\multicolumn{4}{@{}l}{\textit{Judge check: Wan2.2, 30 stories}}\\
Qwen3-VL & \ci{+0.133}{0.044} & \ci{-0.057}{0.018} & \ci{-0.114}{0.034} \\
InternVL3.5 & \ci{+0.123}{0.049} & \ci{-0.028}{0.023} & \ci{-0.050}{0.032} \\
Cross-judge support & $4/4\ \uparrow$ & $3/4\ \downarrow$ & $4/4\ \downarrow$ \\
\midrule
\multicolumn{4}{@{}l}{\textit{Backend check: Qwen3-VL, 8 matched stories}}\\
Wan2.2 & \ci{+0.150}{0.036} & \ci{-0.061}{0.032} & \ci{-0.130}{0.068} \\
MiniMax-H3 & \ci{+0.152}{0.080} & \ci{-0.046}{0.032} & \ci{-0.102}{0.065} \\
\bottomrule
\end{tabular}
\end{table}

\begin{figure}[!t]
\centering
\includegraphics[width=\linewidth]{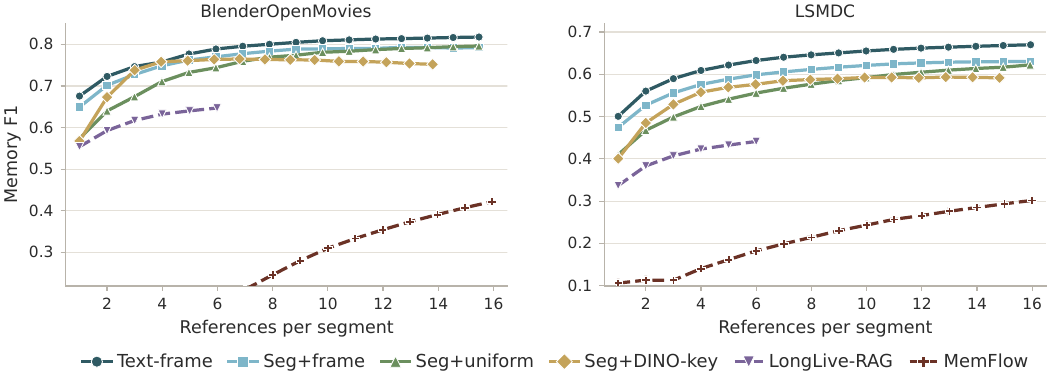}
\caption{\textbf{Track-A reference F1 versus budget.} Each curve keeps a memory's own top-ranked references up to a growing cap. F1 keeps rising with the cap but with shrinking returns, and the ordering holds at every budget: Text-frame leads throughout, and the released memories stay far below. Panels: \dblender{} (left) and \dlsmdc{} (right).}
\label{fig:pareto}
\end{figure}

\begin{figure}[t]
\centering
\includegraphics[width=\linewidth]{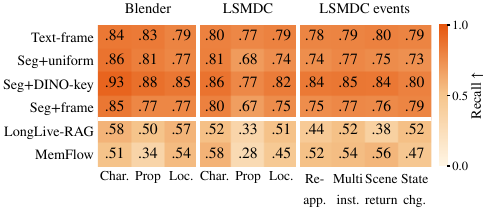}
\caption{\textbf{What is hard on Track-A.} Continuity recall by entity type (first six columns) and LSMDC event (last four). Darker orange means higher recall on a fixed 0--1 scale. Released memories, below the horizontal line, have lower recall across displayed slices. DINO-key leads nine slices; Text-frame leads LSMDC props.}
\label{fig:tracka-difficulty}
\end{figure}

Track-A complements these output measures by scoring retrieved
references before generation. Seg+DINO-key yields the highest reference recall
($0.923$ on \dblender{}, $0.849$ on \dlsmdc{}), but lower
precision ($0.579/0.484$) than Text-frame
($0.692/0.630$). Text-frame instead achieves the best F1
($0.817/0.670$; Table~\ref{tab:main-results}).
The diverse-keyframe policy thus has higher coverage but lower
precision; this comparison does not isolate diversity from its other
retrieval choices. Inspect reference evidence and generated outcomes separately.
These unmatched tracks do not measure a transfer function from
retrieval scores to downstream success. The table also reports retrieval redundancy and selection
efficiency. Across the tested reference caps, F1 rises with diminishing
returns and Text-frame retains the highest F1 (Fig.~\ref{fig:pareto}).
Each cap truncates a policy's own reference ranking; this does not isolate
reference-count effects in generated videos. The
heatmap separates entity and event slices (Fig.~\ref{fig:tracka-difficulty});
among the displayed LSMDC events, DINO-key's recall advantage over
Text-frame is smallest for state changes.

Our location results do not show a retrieval failure: over 425 returns,
retrieval raises whole-frame MegaLoc similarity by $+0.15$ to $+0.21$
over no memory (Supplementary Table~\ref{tab:identity-results}). This embedding
measures scene similarity, not location correctness. Separately, viewpoint
and layout changes can challenge appearance-only cross-view matching, a
problem addressed with geometry cues in Group Editing
~\citep{ma2026groupediting}. Geometry-aware location memory is therefore
an untested hypothesis, not an explanation supported by our results.

\subsection{Clip Quality Is Not Entity Correctness}
\label{sec:exp-quality}

The quality control scores all 2,824 clips per shared-generator
condition with six
VBench-style metrics~\citep{huang2023vbench}. No memory has higher
subject and background consistency despite lower entity recall;
Seg+DINO-key instead has higher imaging and aesthetic scores
(Table~\ref{tab:main-quality}). No quality metric correlates positively
with per-clip entity recall above $\rho=0.17$ in any condition.
Within-clip stability can coexist with entity-level failures, so
quality and entity correctness require separate reporting. This does
not establish independence or eliminate all quality confounds.

Table~\ref{tab:main-quality} averages recall over clips with required
entities, whereas
Table~\ref{tab:trackb-results} pools entity decisions. Native-generator
systems are excluded because their generators differ. Full metrics,
conditions, and protocols are in App.~\ref{app:quality-control}.

\subsection{Sensitivity to the Judge and Video Backend}
\label{sec:exp-sensitivity}

The aggregate trade-off largely survives rescoring the same videos
(Table~\ref{tab:main-sensitivity}). Both judges support recall gains and
Avoid-OK losses for all four retrieval policies. Precision losses are
supported for three of four: Text-frame's interval includes zero under
InternVL3.5. On eight matched stories, MiniMax-H3 replaces Wan2.2 while
retaining FLUX.2 keyframes and preserves all three Seg+DINO-key effects.
Eight stories provide limited coverage of generators and rare states.

Absolute rates remain judge-dependent: no-memory Recall is $0.559$
under Qwen3-VL and $0.361$ under InternVL3.5. Required-present agreement
is only 70--75\% ($\kappa=0.43$--$0.52$), versus 88--91\% for the full
roster. Persistent-state gains and losses for indirect references and false
friends are not significant under the second judge. Full results and agreement statistics: App.~\ref{app:robustness}.

\paragraph{Sensitivity to unparsable judgments.}
Removing 129--143 of 2,824 segments (4.6--5.1\%) per comparison where either
arm's judge output is unparsable preserves all four retrieval policies'
trade-offs: Recall $+0.075$ to $+0.149$, Precision $-0.064$ to $-0.041$,
and Avoid-OK $-0.128$ to $-0.073$. All paired 95\% CIs exclude zero
(Supplementary Table~\ref{tab:robustness}a). Retained samples differ by
policy, limiting effect-size comparisons. This check does not rule out
systematic judging errors or bias from nonrandom missingness.

Crop-conditioned H3 changes the input pathway too; its copying failures
are analyzed separately in App.~\ref{app:robustness}.

\FloatBarrier
\section{Conclusion}
\emph{More memory does not necessarily mean better videos.}
In our controlled study, retrieved history improves entity recall
and resemblance to earlier appearances, yet can reintroduce retired
entities and reinforce obsolete states. Detailed prompts can also
make long-gap generation succeed without visual memory.
The practical lesson is to assess three distinct questions:
\emph{what information history adds}, \emph{which temporal state is
valid}, and \emph{what unwanted content accompanies useful evidence}.
Report these alongside perceptual quality; recall or historical
resemblance alone does not establish memory utility.
\bench provides evidence- and generation-stage measurements, with
an evaluation checklist in App.~\ref{app:reader-guide}.
These are assessment principles, not a validated controller or a
universal memory-capacity law.

\FloatBarrier
\clearpage
\section*{LLM Usage Statement}
We used OpenAI Codex to assist with manuscript writing and editing,
as well as code writing and modification for data summarization,
reproducibility checks, and figure and table preparation.
The authors take full responsibility for the final manuscript, code,
analyses, and reported results.
{
  \small
  \bibliographystyle{ieeenat_fullname}
  \bibliography{references}

\begin{thebibliography}{29}
\providecommand{\natexlab}[1]{#1}
\providecommand{\url}[1]{\texttt{#1}}
\expandafter\ifx\csname urlstyle\endcsname\relax
  \providecommand{\doi}[1]{doi: #1}\else
  \providecommand{\doi}{doi: \begingroup \urlstyle{rm}\Url}\fi

\bibitem[An et~al.(2025)An, Jia, Qiu, Zhou, Huang, Liu, Ren, Kahatapitiya, Liu,
  He, Zhang, Xiang, Yang, Belongie, and Xie]{an2025onestory}
Zhaochong An, Menglin Jia, Haonan Qiu, Zijian Zhou, Xiaoke Huang, Zhiheng Liu,
  Weiming Ren, Kumara Kahatapitiya, Ding Liu, Sen He, Chenyang Zhang, Tao
  Xiang, Fanny Yang, Serge Belongie, and Tian Xie.
\newblock Onestory: Coherent multi-shot video generation with adaptive memory.
\newblock \emph{arXiv preprint arXiv:2512.07802}, 2025.

\bibitem[Dou et~al.(2026)Dou, Li, Cui, Zhou, Wang, and Zhu]{dou2026slotmemory}
Weijia Dou, Hui Li, Jiahao Cui, Lei Zhou, Jingdong Wang, and Siyu Zhu.
\newblock Slotmemory: Object-centric kv memory for streaming long-video
  generation.
\newblock \emph{arXiv preprint arXiv:2605.31033}, 2026.

\bibitem[He et~al.(2026)He, Wei, Yang, and Ordonez]{he2026entitybench}
Ruozhen He, Meng Wei, Ziyan Yang, and Vicente Ordonez.
\newblock {EntityBench}: Towards entity-consistent long-range multi-shot video
  generation.
\newblock \emph{arXiv preprint arXiv:2605.15199}, 2026.

\bibitem[Hu et~al.(2026{\natexlab{a}})Hu, Yang, Huang, Han, and
  Chen]{hu2026longliverag}
Qixin Hu, Shuai Yang, Wei Huang, Song Han, and Yukang Chen.
\newblock Longlive-rag: A general retrieval-augmented framework for long video
  generation.
\newblock \emph{arXiv preprint arXiv:2606.02553}, 2026{\natexlab{a}}.

\bibitem[Hu et~al.(2026{\natexlab{b}})Hu, Wang, and
  McAuley]{hu2026memoryagentbench}
Yuanzhe Hu, Yu Wang, and Julian McAuley.
\newblock Evaluating memory in {LLM} agents via incremental multi-turn
  interactions.
\newblock In \emph{International Conference on Learning Representations},
  2026{\natexlab{b}}.

\bibitem[Huang et~al.(2023)Huang, He, Yu, Zhang, Si, Jiang, Zhang, Wu, Jin,
  Chanpaisit, Wang, Chen, Wang, Lin, Qiao, and Liu]{huang2023vbench}
Ziqi Huang, Yinan He, Jiashuo Yu, Fan Zhang, Chenyang Si, Yuming Jiang, Yuanhan
  Zhang, Tianxing Wu, Qingyang Jin, Nattapol Chanpaisit, Yaohui Wang, Xinyuan
  Chen, Limin Wang, Dahua Lin, Yu Qiao, and Ziwei Liu.
\newblock {VBench}: Comprehensive benchmark suite for video generative models.
\newblock \emph{arXiv preprint arXiv:2311.17982}, 2023.

\bibitem[Ji et~al.(2025)Ji, Chen, Yang, Tao, Wan, and Zhao]{ji2025memflow}
Sihui Ji, Xi Chen, Shuai Yang, Xin Tao, Pengfei Wan, and Hengshuang Zhao.
\newblock Memflow: Flowing adaptive memory for consistent and efficient long
  video narratives.
\newblock \emph{arXiv preprint arXiv:2512.14699}, 2025.

\bibitem[Liu et~al.(2026)Liu, Zhang, Jiang, Wang, Liang, Xue, Yi, and
  Liu]{liu2026iamflow}
Jinzhuo Liu, Jiangning Zhang, Wencan Jiang, Yabiao Wang, Dingkang Liang, Zhucun
  Xue, Ran Yi, and Yong Liu.
\newblock Advancing narrative long video generation via training-free
  identity-aware memory.
\newblock \emph{arXiv preprint arXiv:2605.18733}, 2026.

\bibitem[Ma et~al.(2025)Ma, Feng, Hu, Wang, Wang, Zheng, Wang, Wang, He, Wang,
  Zhu, Liu, He, Wang, Li, Li, Han, Guo, Liu, Xu, Zhang, and
  Chen]{ma2025controllablesurvey}
Yue Ma, Kunyu Feng, Zhongyuan Hu, Xinyu Wang, Yucheng Wang, Mingzhe Zheng,
  Bingyuan Wang, Qinghe Wang, Xuanhua He, Hongfa Wang, Chenyang Zhu, Hongyu
  Liu, Yingqing He, Zeyu Wang, Zhifeng Li, Xiu Li, Sirui Han, Yike Guo, Wei
  Liu, Dan Xu, Linfeng Zhang, and Qifeng Chen.
\newblock Controllable video generation: A survey.
\newblock \emph{arXiv preprint arXiv:2507.16869}, 2025.

\bibitem[Ma et~al.(2026)Ma, Wang, Ma, Wang, Zheng, Yang, Li, Zhao, Ying, Yang,
  Liu, and Chen]{ma2026groupediting}
Yue Ma, Xinyu Wang, Qianli Ma, Qinghe Wang, Mingzhe Zheng, Xiangpeng Yang, Hao
  Li, Chongbo Zhao, Jixuan Ying, Harry Yang, Hongyu Liu, and Qifeng Chen.
\newblock Group editing: Edit multiple images in one go.
\newblock In \emph{Proceedings of the IEEE/CVF Conference on Computer Vision
  and Pattern Recognition (CVPR)}, 2026.

\bibitem[Meng et~al.(2026)Meng, Liu, Ouyang, Wang, Cheng, Yu, Wang, Li, Zhu,
  Zeng, Zhu, Shen, Chen, and Qu]{meng2026causalcine}
Yihao Meng, Zichen Liu, Hao Ouyang, Qiuyu Wang, Ka~Leong Cheng, Yue Yu, Hanlin
  Wang, Haobo Li, Jiapeng Zhu, Yanhong Zeng, Xing Zhu, Yujun Shen, Qifeng Chen,
  and Huamin Qu.
\newblock Causalcine: Real-time autoregressive generation for multi-shot video
  narratives.
\newblock \emph{arXiv preprint arXiv:2605.12496}, 2026.

\bibitem[{MiniMax}(2026)]{minimax2026h3}
{MiniMax}.
\newblock Minimax h3: An open model breaking the boundaries between tasks and
  modalities.
\newblock \url{https://www.minimax.io/blog/minimax-h3}, 2026.
\newblock Accessed: 2026-09-20.

\bibitem[{Qwen Team}(2025)]{qwen2025qwen3vl}
{Qwen Team}.
\newblock Qwen3-vl technical report.
\newblock \emph{arXiv preprint arXiv:2511.21631}, 2025.

\bibitem[Ru et~al.(2024)Ru, Qiu, Hu, Zhang, Shi, Chang, Cheng, Wang, Sun, Li,
  Zhang, Wang, Jiang, He, Wang, Liu, Zhang, and Zhang]{ru2024ragchecker}
Dongyu Ru, Lin Qiu, Xiangkun Hu, Tianhang Zhang, Peng Shi, Shuaichen Chang,
  Jiayang Cheng, Cunxiang Wang, Shichao Sun, Huanyu Li, Zizhao Zhang, Binjie
  Wang, Jiarong Jiang, Tong He, Zhiguo Wang, Pengfei Liu, Yue Zhang, and Zheng
  Zhang.
\newblock {RAGChecker}: A fine-grained framework for diagnosing
  retrieval-augmented generation.
\newblock In \emph{Advances in Neural Information Processing Systems}, 2024.

\bibitem[Shi et~al.(2026)Shi, Li, Deng, Xu, Chen, Wang, Hu, and
  Zhang]{shi2026msvbench}
Haoyuan Shi, Yunxin Li, Nanhao Deng, Zhenran Xu, Xinyu Chen, Longyue Wang,
  Baotian Hu, and Min Zhang.
\newblock {MSVBench}: Towards human-level evaluation of multi-shot video
  generation.
\newblock In \emph{Findings of the Association for Computational Linguistics:
  ACL 2026}, 2026.

\bibitem[{Wan Team}(2025)]{wan2025wan}
{Wan Team}.
\newblock Wan: Open and advanced large-scale video generative models.
\newblock \emph{arXiv preprint arXiv:2503.20314}, 2025.

\bibitem[Wang et~al.(2026{\natexlab{a}})Wang, Shi, Li, Zhou, Liu, Ma, and
  Wan]{wang2026subjectanchor}
Xinyu Wang, Huafeng Shi, Zian Li, Yan Zhou, Xiaoqiang Liu, Yue Ma, and Pengfei
  Wan.
\newblock Subjectanchor: Subject-aware memory-to-video for multi-shot
  storytelling.
\newblock \emph{arXiv preprint arXiv:2609.34502}, 2026{\natexlab{a}}.

\bibitem[Wang et~al.(2026{\natexlab{b}})Wang, Zhao, Zhan, and
  Ma]{wang2026liveedit}
Xinyu Wang, Chongbo Zhao, Fangneng Zhan, and Yue Ma.
\newblock Liveedit: Towards real-time diffusion-based streaming video editing.
\newblock \emph{arXiv preprint arXiv:2606.26740}, 2026{\natexlab{b}}.
\newblock ECCV 2026.

\bibitem[Wu et~al.(2025{\natexlab{a}})Wu, Wang, Yu, Zhang, Chang, and
  Yu]{wu2025longmemeval}
Di Wu, Hongwei Wang, Wenhao Yu, Yuwei Zhang, Kai-Wei Chang, and Dong Yu.
\newblock Longmemeval: Benchmarking chat assistants on long-term interactive
  memory.
\newblock In \emph{International Conference on Learning Representations},
  2025{\natexlab{a}}.

\bibitem[Wu et~al.(2025{\natexlab{b}})Wu, Liu, Zhu, Xia, Feng, Wang, Lin, Shen,
  and Shou]{wu2025moviebench}
Weijia Wu, Mingyu Liu, Zeyu Zhu, Xi Xia, Haoen Feng, Wen Wang, Kevin~Qinghong
  Lin, Chunhua Shen, and Mike~Zheng Shou.
\newblock Moviebench: A hierarchical movie level dataset for long video
  generation.
\newblock In \emph{Proceedings of the IEEE/CVF Conference on Computer Vision
  and Pattern Recognition (CVPR)}, pages 28984--28994, 2025{\natexlab{b}}.

\bibitem[Yang et~al.(2025)Yang, Huang, Chu, Xiao, Zhao, Wang, Li, Xie, Chen,
  Lu, Han, and Chen]{yang2025longlive}
Shuai Yang, Wei Huang, Ruihang Chu, Yicheng Xiao, Yuyang Zhao, Xianbang Wang,
  Muyang Li, Enze Xie, Yingcong Chen, Yao Lu, Song Han, and Yukang Chen.
\newblock Longlive: Real-time interactive long video generation.
\newblock \emph{arXiv preprint arXiv:2509.22622}, 2025.

\bibitem[Yi et~al.(2026)Yi, Kim, Cho, Jang, Yun, and Kim]{yi2026worldkv}
Jung Yi, Minjae Kim, Paul~Hyunbin Cho, Wooseok Jang, Sangdoo Yun, and
  Seungryong Kim.
\newblock Worldkv: Efficient world memory with world retrieval and compression.
\newblock \emph{arXiv preprint arXiv:2605.22718}, 2026.

\bibitem[Yu et~al.(2025)Yu, Bai, Qin, Liu, Wang, Wan, Zhang, and
  Liu]{yu2025contextmemory}
Jiwen Yu, Jianhong Bai, Yiran Qin, Quande Liu, Xintao Wang, Pengfei Wan, Di
  Zhang, and Xihui Liu.
\newblock Context as memory: Scene-consistent interactive long video generation
  with memory retrieval.
\newblock \emph{arXiv preprint arXiv:2506.03141}, 2025.

\bibitem[Yuan et~al.(2026)Yuan, Yin, Li, Huang, Yang, and Yuan]{yuan2026helios}
Shenghai Yuan, Yuanyang Yin, Zongjian Li, Xinwei Huang, Xiao Yang, and Li Yuan.
\newblock Helios: Real real-time long video generation model.
\newblock \emph{arXiv preprint arXiv:2603.04379}, 2026.

\bibitem[Zhang et~al.(2025{\natexlab{a}})Zhang, Jiang, Wang, Fang, Zhi, Yan,
  Kang, Lu, and Pan]{zhang2025storymem}
Kaiwen Zhang, Liming Jiang, Angtian Wang, Jacob~Zhiyuan Fang, Tiancheng Zhi,
  Qing Yan, Hao Kang, Xin Lu, and Xingang Pan.
\newblock Storymem: Multi-shot long video storytelling with memory,
  2025{\natexlab{a}}.

\bibitem[Zhang et~al.(2025{\natexlab{b}})Zhang, Cai, Li, Wetzstein, and
  Agrawala]{zhang2025framepack}
Lvmin Zhang, Shengqu Cai, Muyang Li, Gordon Wetzstein, and Maneesh Agrawala.
\newblock Frame context packing and drift prevention in next-frame-prediction
  video diffusion models.
\newblock \emph{arXiv preprint arXiv:2504.12626}, 2025{\natexlab{b}}.

\bibitem[Zhang et~al.(2026)Zhang, Zhang, Huang, Tang, Wang, Dai, Chen, Li, Li,
  Chen, Liu, Li, and Lyu]{zhang2026mbench}
Shengjun Zhang, Zhang Zhang, Simin Huang, Zhenyu Tang, Hanyang Wang, Chensheng
  Dai, Min Chen, Yifan Li, Yuxin Li, Yingjie Chen, Hao Liu, Chen Li, and Jing
  Lyu.
\newblock Mbench: A comprehensive benchmark on memory capability for video
  world models.
\newblock \emph{arXiv preprint arXiv:2606.00793}, 2026.

\bibitem[Zhao et~al.(2026)Zhao, Wu, Gong, Wang, and Zhao]{zhao2026omnimem}
Lin Zhao, Yushu Wu, Yifan Gong, Yanzhi Wang, and Pu Zhao.
\newblock Omnimem: Scalable and adaptive memory retrieval for long video
  generation.
\newblock \emph{arXiv preprint arXiv:2605.30519}, 2026.

\bibitem[Zhou et~al.(2026)Zhou, Du, Xu, Wang, Zhuang, Zhang, Li, Hu, Su, and
  Chen]{zhou2026videomemory}
Jinsong Zhou, Yihua Du, Xinli Xu, Luozhou Wang, Zijie Zhuang, Y. Zhang, Shuaibo
  Li, Xiaojun Hu, Bolan Su, and Ying-cong Chen.
\newblock {VideoMemory}: Toward consistent video generation via memory
  integration.
\newblock \emph{arXiv preprint arXiv:2601.03655}, 2026.

\end{thebibliography}
}
\clearpage
\appendix
\maketitlesupplementary
\section*{Appendix}
\noindent\textbf{Roadmap.}
Data and cohort coverage: App.~\ref{app:sec-benchmark};
systems, controls, and reproduction: App.~\ref{app:sec-probes};
scoring and calibration: App.~\ref{app:scoring};
additional Track-A and Track-B results: Apps.~\ref{app:tracka-more} and~\ref{app:trackb-more};
author audit: App.~\ref{app:human-attribution};
design implications and limitations: App.~\ref{sec:discussion}.

\section{Evaluation Lessons and Reader Checklist}
\label{app:reader-guide}

\paragraph{What to take away.}
The study identifies three questions that a memory evaluation should
keep distinct: whether history supplies information beyond the current
instruction and any baseline context, whether its appearance matches the requested temporal state,
and whether useful references carry unwanted entities. These are
measurement questions, not a sufficient decision rule for reading memory.
For example, a prompt describing a blue coat may still omit the face;
textual detail alone does not establish that visual history is redundant.
Likewise, an older appearance can be correct for an explicit flashback:
validity is relative to the instruction, not simply reference recency.
Table~\ref{tab:reader-checklist} turns the findings into reporting actions.

\begin{table*}[t]
\centering
\caption{\textbf{A checklist for evaluating visual-memory utility.} Actions motivated by existing findings, not new experimental results. The checks distinguish information contribution, temporal validity, and unwanted content while retaining evidence-stage, quality, and uncertainty controls.}
\label{tab:reader-checklist}
\footnotesize
\setlength{\tabcolsep}{3pt}
\begin{tabular}{@{}>{\raggedright\arraybackslash}p{0.17\linewidth}
                >{\raggedright\arraybackslash}p{0.43\linewidth}
                >{\raggedright\arraybackslash}p{0.34\linewidth}@{}}
\toprule
Assessment & What to report or compare & Interpretation boundary \\
\midrule
Information contribution & Include no memory with the same non-memory inputs; record what appearance information the prompt and any baseline context supply. Where possible, contrast matched prompts with and without that information (Table~\ref{tab:nolook}). & Long absence alone does not show memory was needed. Our stripping test does not establish a reliable retrieval benefit or validate a reading gate. \\
Temporal validity & Report description-matching presence and required state alongside previous-appearance similarity; retain detection/visibility conditions (Table~\ref{tab:main-identity}). & Old resemblance can reward an obsolete look. More recent is not always more valid: flashbacks may require an earlier state. \\
Unwanted content & Report precision and forbidden-entity avoidance with recall, using the same story requirements (Table~\ref{tab:trackb-results}; Sec.~\ref{sec:exp-trackb}). & A recall gain can coincide with extra entities. Whole-policy differences do not isolate a reference's causal contribution. \\
Evidence versus use & Inspect selected-reference coverage/precision separately from generated outcomes; use matched instances if relating the two (Table~\ref{tab:main-results}). & Our Track-A and Track-B use different stories; do not infer a per-story correlation or downstream benefit from Track-A alone. \\
Perceptual quality & Retain clip-quality metrics and entity-level outcomes with explicit aggregation and denominators (Table~\ref{tab:main-quality}). & Within-clip stability is not entity correctness. Weak measured correlation does not establish independence. \\
Uncertainty and controls & State rollout/seed counts, resampling units, judge agreement, budget, and whether the generator is shared (Table~\ref{tab:main-sensitivity}). & Story CIs condition on observed rollouts. Judge-sensitive strata, variable-budget Last-seen, and native-generator rows require separate interpretation. \\
\bottomrule
\end{tabular}
\end{table*}

\paragraph{Choose the measurement boundary first.}
Use Track-A to inspect which references a policy supplies from revealed
real-film history, before generation; use Track-B to inspect clips produced
from the prompt stream and each policy's own generated history. Track-A
scores refer to selected reference items and entity coverage; Track-B
scores refer to rendered appearances. The original tracks use different
stories, so neither their absolute scores nor story-level associations
form a paired retrieval-to-generation analysis. To study that relationship,
use matched inputs and retain both reference and output measurements.

\paragraph{Read outcomes before rankings.}
A description-matching presence hit need not establish the same individual;
a high similarity to the previous appearance need not establish the
correct current state. Similarity conditions on detection in both
appearances, so inspect visibility and missing returns before treating
it as an unconditional success rate. In the shared-generator block,
compare each policy with no memory using the recall/precision/Avoid-OK
vector. Higher values are better for those three rates, but higher
old-appearance similarity is ambiguous after a requested state change.
Native-generator rows compare whole systems; Last-seen is a privileged,
variable-budget diagnostic control. Neither supplies a memory-only ranking
against the shared-generator retrieval policies.

\paragraph{Reproduce the intended object.}
For table inspection, use the frozen analysis JSON and renderers; for
statistical reconstruction, use saved per-segment scores; for a new system,
provide chronological outputs and the correctly indexed scoring view.
App.~\ref{app:reproduction} gives the cohort, code, indexing, and snapshot
checklist. A new rollout is a new measured trajectory, not a byte-identical
reconstruction of the old pixels. Decisions to read, filter, or reject
history remain hypotheses in Table~\ref{tab:directions}; evidence for a
measurement problem alone does not validate its proposed remedy.

\FloatBarrier
\section{Data Construction and Cohort Coverage}
\label{app:sec-benchmark}

\subsection{Benchmark Construction Pipeline (S1--S4)}
\label{app:bench-making}
Track-A gold is built in four stages followed by a deterministic freeze. Model outputs are replaceable
proposals; stable identifiers, prompt completeness, and frozen gold are fixed by deterministic
contracts and human adjudication. \textbf{S1} drafts bounded segments, present entities, first
appearances, and finite-ontology state events; a human-confirmed canonical roster is the naming
authority. \textbf{S2} deterministically normalizes schema, identifiers, timestamps, and duplicates and
runs structural linting. \textbf{S3} reviews each segment with Qwen3-VL-8B~\citep{qwen2025qwen3vl};
low-confidence/conflicting presence claims are rechecked and confirmed entities trigger deterministic
prompt repair. Each unit receives \textsc{Pass}, \textsc{Warn}, \textsc{Block}, or
\textsc{RetryableError}. \textbf{S4} sends every \textsc{Block} plus a stratified audit sample of the
remaining units to blocking human review. Accepted decisions re-run the full deterministic validator.
The final freeze writes the gold registry and layout hash; a changed layout/exclusion set creates a new
data version. Every stage writes a derived artifact and never mutates the original draft.
Fig.~\ref{fig:app-construction} summarizes both tracks' construction, and
Fig.~\ref{fig:app-tracka-draft-gold} shows what the Track-A stages change in practice.

\subsection{Anatomy of a Frozen Gold Instance}
\label{app:bench-data}
\begin{figure}[t]
\centering
\includegraphics[width=\linewidth]{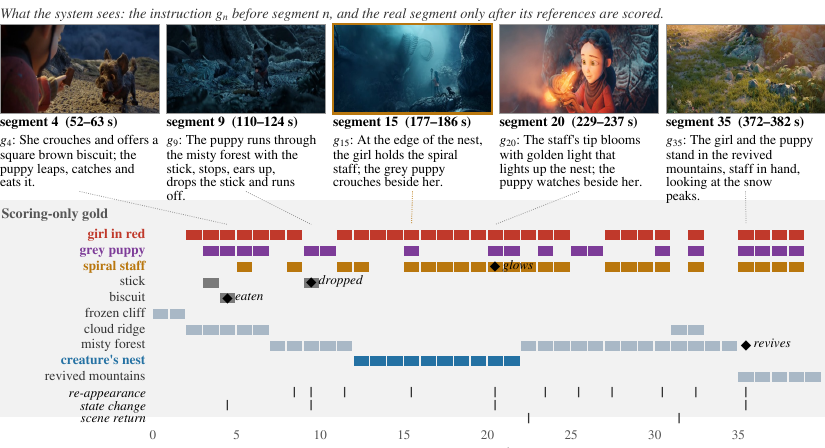}
\caption{\textbf{Anatomy of a Track-A film} (\emph{Spring}, Blender Studio, CC BY 4.0). Top: five story
segments with the instruction $g_n$ the system receives (English excerpts of the Chinese gold). Bottom, on
grey: scoring-only gold for ten of the film's eleven entities over segments 0--39 (the rest are titles and
credits): where each entity is on screen, its state events, and the per-segment diagnostic tags. The grey
puppy leaves after segment 10 and returns in segment 15, the case shown in Fig.~\ref{fig:app-tracka-refs}.}
\label{fig:app-tracka-anatomy}
\end{figure}

A frozen Track-A movie is an ordered sequence of prompt-complete intents with harness-side typed entity
records, lifecycle/state events, required memory records, and forbidden/deprecated records. Only the
current intent and, after reveal, the real segment are visible to the system. Typed gold identities,
persistent gold records, required targets, and forbidden targets remain scoring-only. Entity records
use $\{\texttt{character},\texttt{prop},\texttt{location}\}$ with canonical names and aliases. Track-A's
reappearance, multi-instance, scene-return, and state-change labels are diagnostic tags over this same
frozen structure. Fig.~\ref{fig:app-tracka-anatomy} shows one frozen film.

\subsection{Authored Stories and Cohort Coverage}
\label{app:story-cohort}
\begin{figure}[t]
\centering
\includegraphics[width=\linewidth]{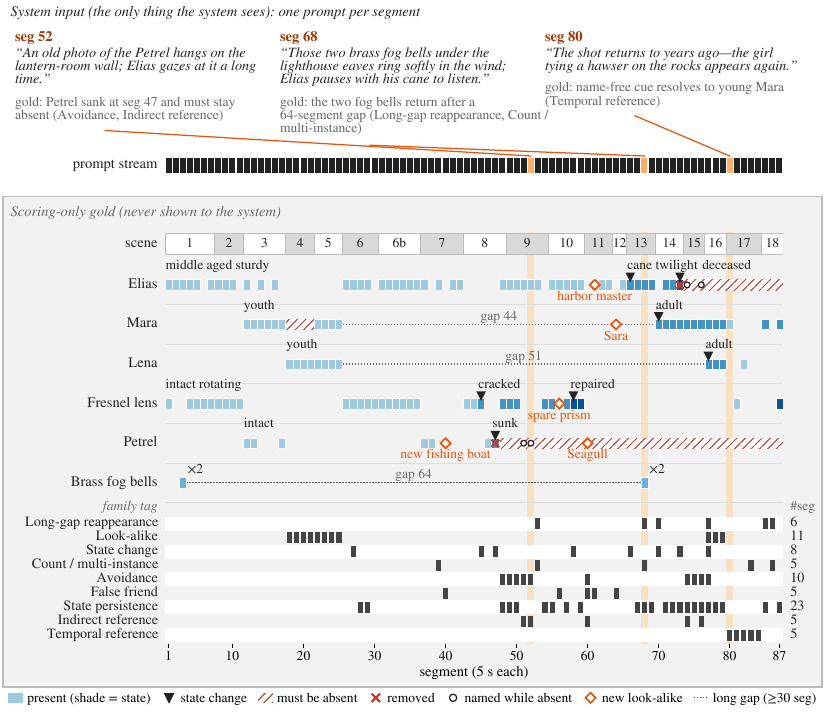}
\caption{\textbf{How a Track-B script plants memory stresses.}
Story \texttt{0001\_lighthouse\_keeper} (87 five-second segments, 19 scenes) drawn as a timeline. The system
receives only the prompt stream (top; three prompts quoted verbatim); everything in the grey panel is
scoring-only gold. Each lane is one of the story's 19 entities: shade gives its authored state, hatching marks
segments where it must be absent (after it is destroyed or dies, or while only its look-alike twin is on
screen), open circles mark prompts that still name a removed entity, and diamonds mark newly introduced
look-alikes; the bottom rows tick the segments tagged with each family of Table~\ref{tab:trackb-families},
with per-family counts at right.}
\label{fig:app-trackb-story}
\end{figure}

A Track-B story is authored rather than annotated: a human writes the entities, their states, look-alike
pairs, and scenes, and the scoring gold and prompt stream are derived from that skeleton.
Fig.~\ref{fig:app-trackb-story} shows how one story places every memory stress on its timeline.

\paragraph{Coverage of the evaluated stories.}
Table~\ref{tab:story-coverage} summarizes only the 30 evaluated stories,
not the larger authoring pool. They contain 2,824 segments, with median
87.5 and range 68--131 segments, and 181 entities with multiple defined
appearance states. Every story contains all nine memory stresses, but
counts are uneven: state-change tags occur in 241 segments and persistent
states in 774. Tags can overlap; these are authored opportunities, not
observed generation outcomes or the scorer's entity-level denominators.
Story titles span settings such as a lighthouse, night market, greenhouse,
and subway, but no genre labels or distribution matching to real films
are established. Topic variety therefore does not establish representativeness.
\begin{table*}[t]
\centering
\caption{\textbf{Coverage of the evaluated Track-B cohort.} Only the 30 stories used in all six seed-2026 conditions are counted. Entity counts are story-local records. Right: tagged segments, number of stories containing the tag, and per-story count range. Tags overlap and are not scoring denominators or observed generation success rates.}
\label{tab:story-coverage}
\footnotesize
\begin{minipage}[t]{0.44\textwidth}\centering
\begin{tabular}{@{}lr@{}}\toprule
Cohort property & Value \\ \midrule
Stories / segments & 30 / 2,824 \\
Segments: mean / median & 94.1 / 87.5 \\
Segments: min--max & 68--131 \\
Scripted duration & 3.92 h \\
Character entities & 152 \\
Prop entities & 300 \\
Location entities & 120 \\
Entities with multiple states & 181 \\
\bottomrule\end{tabular}\end{minipage}\hfill
\begin{minipage}[t]{0.54\textwidth}\centering
\begin{tabular}{@{}lrrr@{}}\toprule
Memory stress & Segments & Stories & Range \\ \midrule
Look-alike & 425 & 30 & 11--61 \\
State change & 241 & 30 & 8--9 \\
Persistent state & 774 & 30 & 19--46 \\
Count memory & 156 & 30 & 5--7 \\
False friend & 150 & 30 & 5--5 \\
Deprecation avoidance & 314 & 30 & 8--16 \\
Indirect reference & 183 & 30 & 4--10 \\
Long-gap return & 224 & 30 & 5--11 \\
Temporal reference & 150 & 30 & 5--5 \\
\bottomrule\end{tabular}\end{minipage}
\end{table*}

\section{Systems, Controls, and Reproduction}
\label{app:sec-probes}
\label{app:system-adaptation}

\subsection{Comparison Scope and Neighboring Work}
\paragraph{Relationship to neighboring memory systems.}
Table~\ref{tab:nearby-memory-work} compares the reported designs and
measurements of OneStory, StoryMem, and SubjectAnchor with our study.
They develop trained memory-conditioned generators and evaluate narrative
quality, alignment, or identity. SubjectAnchor also explicitly studies
shot-level inclusion/exclusion, so entity exclusion itself is not our
novelty. Our diagnostic emphasis is the validity of selected historical
evidence for the current state, its interaction with prompt information,
and the distinction between reference quality and whole-policy outcomes.
The table is a comparison of scope, not a common-protocol performance
ranking; our policy effects do not establish the behavior of all three
trained architectures.
\begin{table*}[t]
\centering
\caption{\textbf{Scope of neighboring visual-memory work.} Designs and evaluation goals reported in the cited papers, rather than scores under a shared protocol. SubjectAnchor explicitly includes inclusion/exclusion control; our distinction concerns evidence-stage diagnostics, current-state scoring, and prompt-information sensitivity. The final row describes the controlled retrieval study, not a newly trained generator.}
\label{tab:nearby-memory-work}
\footnotesize
\setlength{\tabcolsep}{3pt}
\renewcommand{\arraystretch}{1.1}
\begin{tabular}{@{}>{\raggedright\arraybackslash}p{0.13\linewidth}
                >{\raggedright\arraybackslash}p{0.27\linewidth}
                >{\raggedright\arraybackslash}p{0.25\linewidth}
                >{\raggedright\arraybackslash}p{0.29\linewidth}@{}}
\toprule
Work & Historical visual context & Learned conditioning & Reported evaluation focus \\
\midrule
OneStory\newline\citep{an2025onestory} & Semantically selected frames from earlier shots; compact global memory. & Fine-tuned I2V model with frame selection and an adaptive patch conditioner. & Character/environment consistency, shot quality, and semantic alignment in text- and image-conditioned multi-shot generation. \\
StoryMem\newline\citep{zhang2025storymem} & Generated-shot keyframes selected semantically and filtered by aesthetic preference; dynamic memory bank. & LoRA-trained memory-to-video model using latent concatenation and negative temporal RoPE shifts. & ST-Bench: cross-shot consistency, global/single-shot prompt following, aesthetic quality, and human preference. \\
SubjectAnchor\newline\citep{wang2026subjectanchor} & Subject-related precomputed keyframes traced to historical appearances. & Trained Wan2.2-I2V with subject-aware RoPE and partitioned memory attention. & Subject/prompt alignment, identity relative to first/previous appearances, visual quality, and inclusion/exclusion control. \\
\bench\newline(this study) & Real-film evidence replay (Track-A); references from each policy's generated history (Track-B). & Shared FLUX.2/Wan2.2 pipeline across retrieval policies and controls; no newly trained memory architecture. & Reference coverage/precision before generation; required and forbidden appearances, current state, conditional appearance similarity, prompt stripping, and judge/backend sensitivity. \\
\bottomrule
\end{tabular}
\end{table*}

\paragraph{SubjectAnchor comparison feasibility.}
SubjectAnchor uses a subject-aware trained architecture, a 50-story
benchmark, and precomputed subject-centric keyframes with Wan2.2-I2V
\citep{wang2026subjectanchor}. Our shared-generator comparison uses
FLUX.2 keyframe composition followed by Wan2.2-TI2V. The public preprint
links no implementation or checkpoint for running its model under our
protocol, so a matched empirical comparison is not currently
reproducible. We compare its stated task and findings, without claiming
that our observed trade-off applies to every memory architecture.

\subsection{Causal Adapter Contract}
Released causal systems are driven through authors' released memory mechanisms and weights with only
protocol-required substitutions; the four retrieval memories are ours and form the controlled testbed. All scored
rows share frozen gold, causal guard, materialization, and scoring. No gold crop or gold identity is
exposed to an ordinary system; only the privileged last-seen control uses gold IDs
for reference selection (App.~\ref{app:last-seen}). On Track-A, self-generated observation is replaced by the revealed real segment; if
a native write requires a model forward, we use one teacher-forced clean-context forward without
denoising. Recalled items are mapped back to source frames through temporal identities and dropped if
they violate the causal guard. Methods that require future story information or modalities absent from
passive film are not assigned surrogate quantitative scores. Table~\ref{tab:memory-interfaces} summarizes
why the families' internal diagnostics are not comparable and which common observables \bench uses.

\begin{table}[!t]
\centering
\caption{{\normalfont Different internal memories, one common interface.}
The families expose different internal states and native diagnostics, so those quantities are not directly comparable and do not by themselves establish what reaches or survives generation. \bench{} maps them to common observables: boundary context $\widehat{\mathcal{C}}_n$ before synthesis (Track-A) and generated outcome $x_n$ after synthesis (Track-B).}
\label{tab:memory-interfaces}
\footnotesize
\setlength{\tabcolsep}{3pt}
\renewcommand{\arraystretch}{1.08}
\begin{tabular}{@{}
  >{\raggedright\arraybackslash}p{0.13\linewidth}
  >{\raggedright\arraybackslash}p{0.33\linewidth}
  >{\raggedright\arraybackslash}p{0.44\linewidth}
  @{} }
\toprule
{\normalfont\bfseries Memory family}
  & {\normalfont\bfseries Internal interface / native diagnostic}
  & {\normalfont\bfseries What internal-only evaluation cannot establish} \\
\midrule
Window / KV
  & frames, latents, or KV; recency / native attention; cache reach
  & Retention does not imply that intent-relevant evidence crosses the finite conditioning boundary; stale history may remain active. \\
Retrieval
  & indexed frames / clips; search $+$ rank; Recall@$K$ / similarity
  & Larger $K$ can buy recall while increasing context spend or contamination; a good retrieval decision may still fail in synthesis. \\
Entity / state memory
  & persistent identity / state records; address, update, filter; record validity
  & Correct records do not guarantee that the right visual evidence is materialized under budget or realized correctly by the generator. \\
\bottomrule
\end{tabular}
\end{table}

Table~\ref{tab:memory-read-forget} lists, for each memory we cite or test, what it stores, how it decides
what to read, and whether and how it forgets.

\begin{table*}[t]
\centering
\caption{\textbf{How causal video memories read and forget.}
Every read matches the current instruction, latest frame, camera pose, or attention query by similarity or relevance, and forgetting is either absent, a budget eviction ranked by that same relevance, or age-based compression; no causal system evicts or withholds an entry because the story changed its state. Only IAMFlow and the non-causal VideoMemory keep explicit entity records, and both address them by the entities named in the current instruction or shot.}
\label{tab:memory-read-forget}
\footnotesize
\setlength{\tabcolsep}{4pt}
\renewcommand{\arraystretch}{1.08}
\begin{tabular}{@{}
  >{\raggedright\arraybackslash}p{0.16\linewidth}
  >{\raggedright\arraybackslash}p{0.21\linewidth}
  >{\raggedright\arraybackslash}p{0.24\linewidth}
  >{\raggedright\arraybackslash}p{0.21\linewidth}
  >{\raggedright\arraybackslash}p{0.08\linewidth}
  @{} }
\toprule
{\normalfont\bfseries System}
  & {\normalfont\bfseries Stored as}
  & {\normalfont\bfseries Read by}
  & {\normalfont\bfseries Forgets by}
  & {\normalfont\bfseries Causal} \\
\midrule
\multicolumn{5}{@{}l}{\textit{Released / published memories}} \\
FramePack \citep{zhang2025framepack}
  & Frame latents, compressed by importance rank
  & All history, ordered by recency or similarity
  & Geometric compression; tail deleted or pooled
  & Partly (endpoint plans) \\
MemFlow \citep{ji2025memflow}
  & Local KV window $+$ KV bank
  & Prompt-to-frame attention, top-$k$
  & Budget; keeps top-$k$ by prompt relevance
  & Yes \\
IAMFlow \citep{liu2026iamflow}
  & Entity-ID registry $+$ archived frame KV
  & Set cover of IDs named in prompt
  & None; archive grows, read capped
  & Yes \\
LongLive-RAG \citep{hu2026longliverag}
  & Evicted latent KV $+$ AE embedding
  & Cosine to latest latent, top-$K$
  & None; every evicted frame kept
  & Yes \\
Context-as-Memory \citep{yu2025contextmemory}
  & Raw generated frames
  & Camera FOV overlap with target view
  & None; read capped at $k$
  & Yes \\
WorldKV \citep{yi2026worldkv}
  & Evicted KV chunks, pose-indexed
  & Camera/action pose distance, top-$k$
  & No chunk eviction; redundant tokens pruned
  & Yes \\
OmniMem \citep{zhao2026omnimem}
  & Full-history KV blocks, GPU/CPU
  & Per-head attention top-$K$, local excluded
  & None; LRU only offloads to CPU
  & Yes \\
SlotMemory \citep{dou2026slotmemory}
  & Slot-indexed KV tokens, fixed bank
  & Prompt-text $+$ visual relevance, top-$k$
  & Budget; evicts least prompt/visual relevance
  & Yes \\
VideoMemory \citep{zhou2026videomemory}
  & Entity ID, attribute text, reference image
  & LLM attribute match per planned shot
  & None; new states appended
  & No (full script) \\
\midrule
\multicolumn{5}{@{}l}{\textit{Our testbed probes (App.~\ref{app:retrieval-probes})}} \\
Text-frame
  & Segment text $+$ timestamped frames
  & Instruction-to-frame cross-modal similarity
  & None; append-only, read capped
  & Yes \\
Seg+uniform
  & Segment text $+$ timestamped frames
  & Segment text similarity; uniform frames
  & None; append-only, read capped
  & Yes \\
Seg+DINO-key
  & Segment text $+$ timestamped frames
  & Segment text similarity; DINOv3 diverse frames
  & None; append-only, read capped
  & Yes \\
Seg+frame
  & Segment text $+$ timestamped frames
  & Segment text, then frame rerank (RRF)
  & None; append-only, read capped
  & Yes \\
\bottomrule
\end{tabular}
\end{table*}

\subsection{The Four Retrieval Memories}
\label{app:retrieval-probes}
\paragraph{Track-A retrieval definitions.}
The adapter stores observed segment text and sampled video frames with temporal identities.
\textbf{Text-frame} ranks historical frames by cross-modal similarity. \textbf{Seg+uniform} ranks
segments by text similarity and uniformly samples frames. \textbf{Seg+DINO-key} uses the same
segment ranking, 2-fps dense sampling, and greedy farthest-point DINOv3 diversity. \textbf{Seg+frame
reranking} performs text coarse retrieval followed by frame-level cross-modal reranking with reciprocal
rank fusion. Track-A reports their native cap $B{=}16$ and the budget sweep in Fig.~\ref{fig:pareto}.

\paragraph{Audited Track-B generation loop.}
All six shared-generator conditions (the four retrieval memories and the two controls) come from one run of
the released testbed (\texttt{vmem\_bench.testbed}, seed 2026): one rollout per story and condition, all
generated with the same code revision and scored by the same judge. Each story starts with an empty retrieval store and an empty entity bank.
At segment $n$, the authored prompt $g_n$ is used directly as the retrieval query. The history record for
each prior segment stores that same authored prompt as both prompt text and description, together with
frames decoded from that memory's own previously generated MP4; no real-film frame, gold crop, or other
memory's trajectory enters the archive. Retrieval happens before write-back, with eligibility requiring the
historical segment end time to be no later than the store clock. The resulting still images are then passed
as composed references to a shared keyframe-first stack: FLUX.2 Klein 9B-KV FP8 composes either a text-only
or multi-reference keyframe, and Wan2.2-TI2V-5B-Turbo expands it to a 121-frame segment (1280$\times$704
render, delivered at 832$\times$480). The generated segment is then appended to that memory's own store and
becomes available to later segments. The retrieval reference budget is $K{=}3$; the audited code defaults are
1-fps uniform sampling, 2-fps dense sampling, top-5 text-ranked segments, three DINO keyframes per segment,
and reciprocal-rank fusion with $k{=}60$ for Seg+frame reranking. The scorer is the same pinned Qwen3-VL-32B judge used for every Track-B row, applied to the frozen story gold.

\paragraph{Realization-layer scope.}
The reported retrieval rollouts are keyframe-first: selected stills are composed into a single FLUX
scene keyframe, and Wan receives that image as its first-frame condition. This is useful for holding a
simple realization mechanism fixed, but it introduces a structural bias: memory that should emerge only
later in the segment may be pulled into the first frame. The codebase also contains Wan2.2-I2V-A14B
backends with SVI/Morphic LoRA variants, but the reported Track-B retrieval rows did \emph{not} use those
paths; they used Wan2.2-TI2V-5B-Turbo. A unified multimodal reference-conditioned generator
such as MiniMax H3~\citep{minimax2026h3}, which accepts text, image, video, and audio context directly,
is a natural future replacement because it may avoid collapsing multiple memory cues into one forced
first-frame keyframe. When we designed the Track-B realization setup, H3 had not yet been open-sourced;
the main results therefore use the FLUX-to-Wan scaffold. We have since run all six conditions on 8 stories
with two H3 realizations (App.~\ref{app:robustness}, Table~\ref{tab:second-pipeline}); we make no claim
that H3 improves VMem-Bench scores.

\paragraph{Provenance.}
The testbed's production files are internally 0-indexed whereas Track-B gold is 1-indexed, so scoring reads
a view that maps \texttt{GT seg\_k} to production \texttt{seg\_\{k-1\}}. Every condition scores all
2,824/2,824 segments. The release pins the code revision and the scored snapshots rather than a moving
source tree.

\subsection{Last-Seen Reference Control}
\label{app:last-seen}
For Track-B segment $n$, the control reads the gold cast and keeps the characters and props that must appear
and have appeared before. It orders them by gap (longest first), keeps up to three, and for each takes the
middle frame of this rollout's own clip of the segment in which the entity last appeared; duplicates are
merged. The generator receives only these frames, never gold text, names, or images, and a guard admits
only clips that this run produced strictly before segment $n$. Because the needed entity was usually on
screen just before, 1,803 segments receive one reference, 842 two, 149 three, and 30 (the first segment of each
story) none. The control shows what a memory gains by choosing the right entity at its latest moment; it is
not an information upper bound, since it hands over fewer references than retrieval and its frame may
itself be a wrong earlier rendering.

\subsection{Released Causal-Memory Systems}
\paragraph{LongLive-RAG.}
Each revealed Track-A segment is VAE-encoded to Wan latents~\citep{wan2025wan} with the released
LongLive-RAG~\citep{hu2026longliverag} wrapper and published LatentAE. The released latent-memory rule
performs cosine top-$k$ retrieval with \texttt{memory\_size=6}, \texttt{recent\_exclude=5}, and
\texttt{sink\_size=1}. Track-B uses the released generator-coupled system.

\paragraph{MemFlow.}
MemFlow~\citep{ji2025memflow} combines a recent window with a compressed long-term KV bank. We
teacher-force each revealed Track-A segment through the released base+LoRA clean-context commit path,
record the local window and surviving bank blocks, and recover source frames via bit-exact fingerprints
of retained KV blocks. The temporal RoPE table is extended to the film length with the same closed-form
rotary embedding. Track-B uses the released generator-coupled path.

\paragraph{IAMFlow.}
IAMFlow~\citep{liu2026iamflow} scores frames using the native diffusion transformer's KV cache. Its
offline driver precomputes entities from the full prompt list; our causal adapter disables this future
look-ahead and uses native per-prompt processing. Track-B uses the released generator-coupled stack.
Track-A is reported only as a scalability diagnostic because the native per-frame KV archive exceeds
host memory on long films.

\subsection{IAMFlow Track-A Scalability}
\label{app:iamflow-scalability}
IAMFlow's native long-term archive copies per-frame K/V slices to host memory. For its
Wan2.1-T2V-1.3B backbone, one archived frame costs approximately
$30\times2\times1560\times1536\times2$ bytes $\approx274$ MiB. In our driver, 15-second segments add
about 5.3 GiB of anonymous host memory per segment; with an 800--900 GiB host-memory budget, the native
carrier reaches a hard horizon of roughly three thousand archived KV frames.
\begin{table*}[t]
\centering
\caption{IAMFlow host-memory growth observed during our Track-A scalability test. The limiting
quantity is the number of archived KV frames, which scales with total observed video duration rather
than simply with the number of benchmark segments. The predicted GiB column is the archived-KV tensor
payload from shape alone; OOM rows exceed the fixed host-memory budget before a full-movie scored
output can be produced.}
\label{tab:iamflow-host-memory}
\small
\setlength{\tabcolsep}{4pt}
\begin{tabular}{@{}lrrrrl@{}}
\toprule
Movie & Segs. & Mean seg. & KV frames & Pred. GiB & Outcome \\
\midrule
\texttt{0053\_Rendezvous\_mit\_Joe\_Black} & 221 & 7.3\,s  & 2071 & 555  & completed \\
\texttt{sita\_sings\_the\_blues\_part1}    & 164 & 14.9\,s & 3259 & 873  & OOM \\
\texttt{0013\_Halloween}                   & 128 & 14.9\,s & 2442 & 654  & completed \\
\texttt{1005\_Signs}                       & 296 & 9.8\,s  & 3763 & 1008 & OOM \\
\texttt{0017\_Pianist}                     & 235 & 13.8\,s & 4287 & 1148 & OOM \\
\bottomrule
\end{tabular}
\end{table*}

We do not substitute a bounded-KV variant because it changes the memory carrier used for recall.

\subsection{Reproduction Checklist}
\label{app:reproduction}
The current shared-generator results use run tag
\texttt{testbed\_rerun\_20260927}, seed 2026, and code revision
\texttt{0b41d3d}; the older retrieval batches and prospective three-seed
commands do not describe additional measured seeds. The reproduction
checklist separates three operations: rendering frozen analysis tables,
recomputing statistics from saved scores, and generating new video.

\begin{enumerate}[leftmargin=1.25em,itemsep=1pt,topsep=2pt]
\item Use the exact 30-story cohort and its gold/prompt SHA256 list;
      the larger authoring pool is not the evaluated cohort.
\item Select the six conditions, fixed prompts, seed, and generator
      configuration; keep Last-seen's privileged plan separate.
\item Map gold segment $k$ to generated segment $k-1$ through the
      gold-numbered scoring view before calling the scorer.
\item Pin judge models/settings, retain per-segment scores and parse
      failures, and use the specified paired story bootstrap.
\item Regenerate tables from the frozen analysis JSON, without replacing
      those inputs with newly scored or partially completed runs.
\item Check artifact distribution separately: local result symlinks do
      not establish that a public snapshot is downloadable.
\end{enumerate}
A metadata audit found 180 saved run manifests from the same clean
revision, each condition recording all 2,824 successful segments, and
180 corresponding score snapshots. Current cohort gold and prompts
match the pinned generation checkout byte for byte. This checks saved
metadata and file fingerprints; it does not regenerate pixels or
verify external download availability. Only the bundled example is
certified by the legacy freeze manifest; the cohort's 60 input hashes
are recorded in the separate coverage snapshot.

\FloatBarrier
\section{Scoring Protocols and Calibration}
\label{app:scoring}

\begin{table*}[t]
\centering
\caption{\textbf{Calibration of the identity-consistency metrics.} Positive controls pair two disjoint frame sets of the same clip; negative controls pair two different entities of the same kind and story. For the embeddings, the threshold is the Youden-optimal cut on these controls, fixed before any condition was scored. The VLM judge answers the same question twice with 0.993--1.000 exact agreement.}
\label{tab:identity-calibration}
\footnotesize
\setlength{\tabcolsep}{4pt}
\begin{tabular}{@{}l r r r r r@{}}
\toprule
Metric & pos./neg. pairs & AUC & threshold & pos. median & neg. median \\
\midrule
DINOv3 crop & 385/389 & 1.000 & 0.67 & 0.96 & 0.19 \\
ArcFace face & 201/157 & 0.997 & 0.35 & 0.84 & 0.11 \\
MegaLoc scene & 200/200 & 1.000 & 0.66 & 0.89 & 0.07 \\
DINOv3 scene & 200/200 & 1.000 & 0.84 & 0.98 & 0.17 \\
\midrule
VLM pairwise judge & \multicolumn{5}{l}{same when both visible: 0.98 (pos.), 0.78 (neg.); neg.\ not consistent: 0.93} \\
\bottomrule
\end{tabular}
\end{table*}

\subsection{Entity Scoring and Statistical Uncertainty}
\label{app:bench-scoring}
\paragraph{Track-A metrics.}
Precision, Recall, F1, $\mathrm{SelEff}$, $R_V$, and $R_S$ are defined in
Sec.~\ref{sec:bench-metrics}. The judge scores each reference independently against the frozen roster
and the real next segment. Redundancy then uses multimembership groups, so a reference that covers
several gold entities contributes to each of those groups rather than to an arbitrary first label.
$R_V$ is repeated-view redundancy inside those groups, and $R_S$ is the pair-weighted mean DINOv3
cosine among crops of the same entity. Headline Precision, Recall, and F1 are not rewritten by that
grouping. Dataset numbers are unweighted means over a film's segments, then over films.

\paragraph{Causal guard and budget sweeps.}
Every reference carries source time and is materialized at that recalled time. Any item at or after the
current segment start is dropped. Fig.~\ref{fig:pareto} truncates each system's own ranked output at
increasing caps $K=1,\ldots,16$; it is a diagnostic budget sweep, not per-$K$ retuning.

\paragraph{Track-B metrics.}
Each family $f$ defines an opportunity set $\mathcal{O}_f$ and predicate $\mathrm{sat}_f(o)$ with
$p_f=|\mathcal{O}_f|^{-1}\sum_{o\in\mathcal{O}_f}\mathrm{sat}_f(o)$. Must-remember families use recall,
identity/state/count families use accuracy, and must-not-render families use avoidance-OK.

Family definitions and native metrics are in Table~\ref{tab:trackb-families}.

\begin{figure}[t]
\centering
\includegraphics[width=\linewidth]{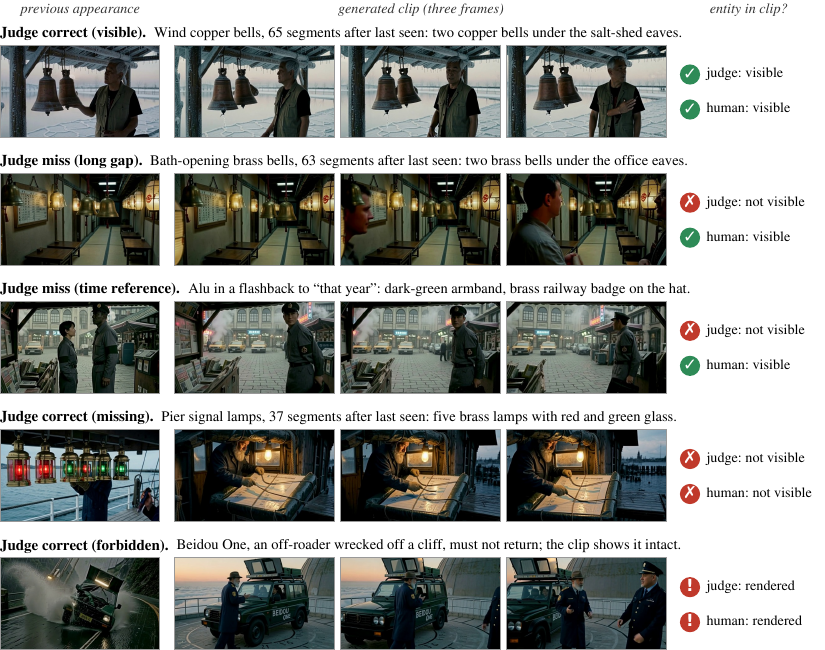}
\caption{\textbf{Track-B judge verdicts checked against human labels on real units.} Each row shows the
entity's previous appearance in the same rollout, three frames of the generated clip, the verdict of the
pinned Qwen3-VL-32B judge, and the human label for the clip. In rows~1, 4 and~5 the judge is right about a
required entity being present or absent and about a forbidden entity being rendered; 28 of 32
judge-confirmed controls were confirmed. In rows~2 and~3 the judge reports a required entity as missing
although it is clearly rendered; such misses make up 27\% of sampled required-entity failures and
concentrate at long gaps and time references, so Track-B recall there is conservative.}
\label{fig:app-judge-examples}
\end{figure}

Track-B uses one deterministic judge call per segment ($k_{\mathrm{votes}}{=}1$, temperature $0$). The
blinded roster mixes present, forbidden/deprecated, look-alike, and deterministic decoy entities. Decoy
FPR / abstention for MemFlow, IAMFlow, and LongLive-RAG are $0.050/0.057$, $0.054/0.049$, and
$0.016/0.132$, respectively; the last pair illustrates why precision-like metrics must be read with
coverage. When the judge returns no parsable answer for a segment, every entity on its list counts as
abstained: required entities are missed and forbidden ones absent. This happens for 91, 101, 112, and 115 of
the 2,824 segments of Text-frame, Seg+uniform, Seg+DINO-key, and Seg+frame, and for 55 of the no-memory
control (2--4\%), so it slightly lowers the retrieval memories' recall relative to the control.
Fig.~\ref{fig:app-judge-examples} shows judge verdicts next to human labels on real units, including
the two kinds of miss that make Track-B recall conservative.

\paragraph{Uncertainty of memory effects.}
Every difference between a retrieval memory and the no-memory control is given with a paired,
story-clustered bootstrap: the 30 stories are resampled with replacement 2,000 times (seed 20260925), the
same draw is applied to both conditions, and counts are pooled within the draw before taking the rate. The
95\% interval is the 2.5th--97.5th percentile of the difference; all values are in
\path{tables/experiments/memory_gain_bootstrap.json}. Each condition is one closed-loop rollout per
story, so these intervals capture story-level variation, not rollout-to-rollout variation.

\subsection{Identity Measurement and Calibration}
\label{app:identity}
The presence judge checks each entity against its gold description only, so it cannot tell a returning
character from a description-matching stranger. Two versioned scorers in \bench measure identity directly;
the full protocol, prompts, thresholds, and model installer are in the benchmark's scoring documentation.

\paragraph{Units.} From the frozen gold, one unit per return of a character or prop that was absent in the
previous segment but seen before (1,130 units), plus a seeded 20\% sample of returns from the previous
segment (583) and the characters that return changed after 30 or more segments (61). The previous
appearance is the latest earlier segment whose cast lists the entity; both appearances come from the same
rollout of the same condition.

\paragraph{VLM pairwise judge (\texttt{trackB-identity-1.0}).} For each appearance, three frames (15, 50, and 85\%
of the clip) side by side. The judge receives both strips, the entity's name, kind, and gold description at
each appearance (with a note when a state change is expected), is told to ignore camera, pose, lighting, and
background, and answers in JSON whether the entity is visible in each strip and, if in both, whether it is the
same individual (face, hair, build, clothing; shape, color, material, markings for props). The condition is
hidden. The scorer also reports the share of units where the entity is visible in both clips and judged the
same, but this rate depends on how well the \emph{previous} appearance was rendered, which memory affects
through the earlier segments of the rollout: for characters returning changed after 30 or more segments,
the current clip shows the character in almost every condition ($0.98$--$1.00$) and visible pairs are
almost always called the same ($0.93$--$1.00$), so nearly the whole difference between conditions ($0.10$ vs.\
$0.46$--$0.57$) comes from the old appearance. We therefore report the same-individual rate among returns visible in both clips, which
compares the two appearances only.

\paragraph{Embeddings (\texttt{trackB-embed-1.0}).} GroundingDINO localizes the entity in each frame from its
name and description; the crop is embedded with DINOv3 and, for characters, the largest face in the crop with
ArcFace. Locations use the whole frame through MegaLoc, a place-recognition model, with whole-frame DINOv3 as
a secondary check; whole-frame similarity is not used for characters or props, since a memory that copies a
reference's composition would score high without the entity being the same. The similarity of a unit is the
highest cosine over frame pairs detected on both sides; we report its mean over units detected on both sides.

\paragraph{Calibration and uncertainty.} Positive controls pair two disjoint frame sets of one clip; negative
controls pair two different entities of the same kind and story (Table~\ref{tab:identity-calibration}). The
embeddings separate them almost perfectly, the VLM judge less so: it rarely calls two visible look-alikes
different, so its same-individual rate is optimistic. DINOv3 crop similarity is therefore the primary
identity measure and the VLM judge a check.
Three frames can miss a brief appearance. Differences to no memory use the same paired story bootstrap as
all Track-B results (Table~\ref{tab:identity-results}).

\FloatBarrier
\section{Additional Track-A Results}
\label{app:tracka-more}

\subsection{Coverage, Precision, and Redundancy}
\begin{figure}[t]
\centering
\includegraphics[width=\linewidth]{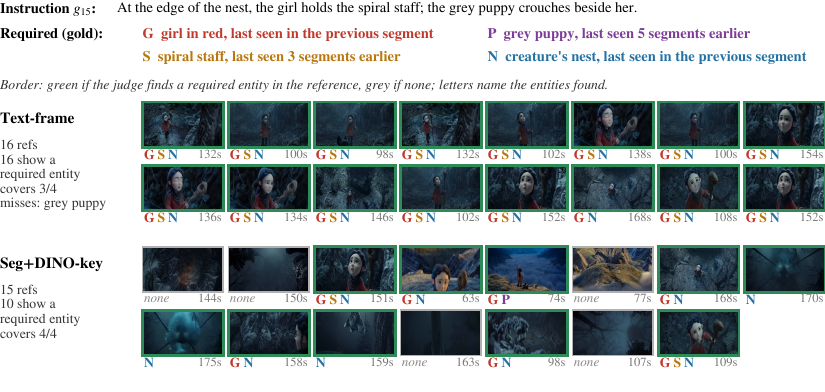}
\caption{\textbf{Every reference two memories hand over for \emph{Spring} segment 15.} Text-frame returns 16
near-identical views of the girl in the forest and never the puppy, last seen five segments earlier.
Seg+DINO-key samples diverse views, finds the puppy on the cloud ridge (74\,s), and covers all four entities,
but five of its 15 references show none of them. Letters are the automatic judge's labels; it also tags some
misty-forest frames before 144\,s as the nest, which does not change which references count as correct here.}
\label{fig:app-tracka-refs}
\end{figure}

The repetition columns of Table~\ref{tab:main-results} separate two kinds of redundancy. Seg+DINO-key has
the best selection efficiency ($0.431/0.393$), the lowest appearance similarity $R_S$ ($0.488/0.393$), and
near-lowest view repetition $R_V$ ($0.225/0.143$; MemFlow $0.133$ on \dlsmdc{}), yet it is the least precise
retrieval memory: diverse views are not the same as correct views. Fig.~\ref{fig:app-tracka-refs} shows
both effects on one segment.

\subsection{Reference Budget} Fig.~\ref{fig:pareto} lets each memory hand over more of its own top-ranked references:
F1 rises with shrinking returns, and the ordering of the memories holds at every tested budget.
The curves are shown in the main paper; they measure reference evidence, not generation utility.

\subsection{Growing History} 
\begin{figure}[ht]
\centering
\includegraphics[width=\linewidth]{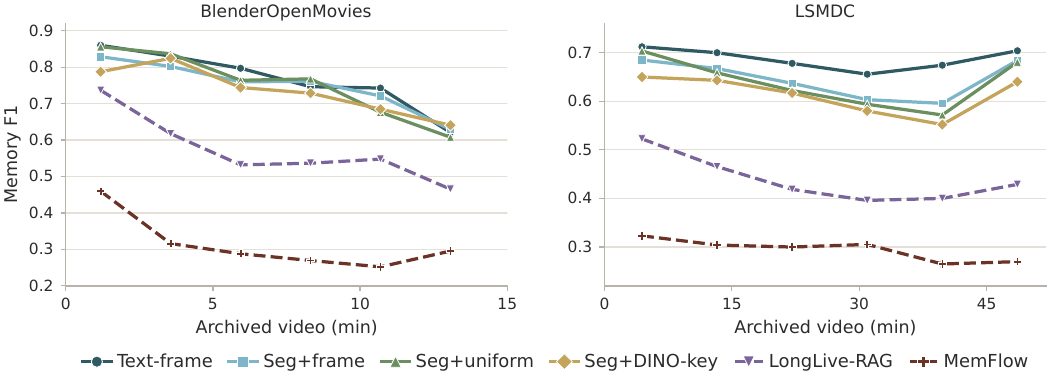}
\caption{\textbf{Track-A memory F1 versus archived-video length} (same runs as
Table~\ref{tab:main-results}). Retrieval remains comparatively stable as history grows; the released
causal-memory paths lose more F1. The far bins contain few films.
Left: \dblender{}; right: \dlsmdc{}.}
\label{fig:tracka-horizon}
\end{figure}

Fig.~\ref{fig:tracka-horizon} tracks F1 as the archived film grows: retrieval
stays nearly flat, while the released memories keep losing; since they evict or compress entries while the
testbed re-ranks the full history, this decline mixes lossy storage with reading.

\FloatBarrier
\section{Additional Track-B Results}
\label{app:trackb-more}

\subsection{Qualitative Rollouts}
\label{app:qualitative-rollouts}
\begin{figure*}[!t]
\centering
\includegraphics[width=.90\textwidth,height=.68\textheight,keepaspectratio]{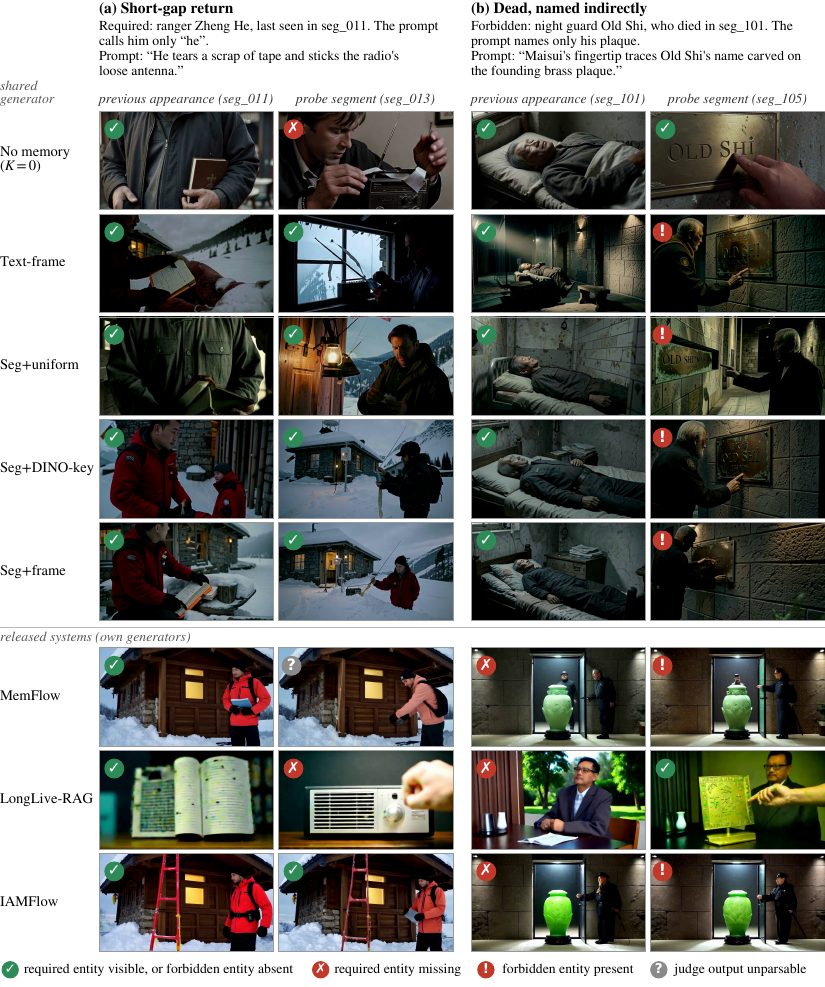}
\caption{\textbf{Track-B rollouts on two memory probes: memory brings back a recently seen character, and
also a dead one.} Rows are systems given the same prompt stream; columns show the entity's previous
appearance and the probe segment (middle frame of each generated clip). In (a) the no-memory generator draws
a stranger for ``he'' while all four retrieval memories restore the ranger; in (b) no memory keeps the dead
guard out, but all four retrieval memories, MemFlow, and IAMFlow put him back at his plaque. Badges are the
automatic judge's verdicts for that entity, and Track-B Recall and Avoid-OK pool such verdicts over all
2,824 segments; the released systems use their own generators.}
\label{fig:app-trackb-rollouts}
\end{figure*}

Fig.~\ref{fig:app-trackb-rollouts} shows what the Track-B numbers look like in pixels (from an earlier
rollout of the same testbed): on the same prompt stream, retrieval restores a character seen one segment earlier that the no-memory generator replaces with
a stranger, and it also returns a dead character whom the prompt names only through his plaque.

\subsection{Return Gaps and Prompt Information} 
\begin{figure}[!t]
\centering
\begin{minipage}[t]{0.49\linewidth}
\centering
\resizebox{\linewidth}{!}{%
\begin{tikzpicture}[baseline=(current bounding box.north)]
\begin{axis}[
  scale only axis,
  width=0.78\linewidth,height=3.15cm,
  ymin=0.10,ymax=0.95,
  xtick={1,2,3},
  xticklabels={Near,Mid,Far},
  ylabel={Continuity recall},
  title={(a) Recall versus absence gap},
  title style={font=\bfseries\scriptsize,anchor=south west,at={(0,1.04)}},
  tick label style={font=\scriptsize},
  label style={font=\scriptsize},
  grid=major,grid style={gray!18},
  legend cell align={left},
  legend columns=2,
  legend to name=trackb-leg-a,
  every axis plot/.append style={line width=0.8pt,mark size=1.5pt},
  legend image post style={xscale=0.6},
  legend style={font=\fontsize{5.4}{6.2}\selectfont,draw=none,fill=none,
    /tikz/every even column/.append style={column sep=3pt},
    row sep=0.5pt},
]
\addplot+[mark=*,color=sysTF,mark options={fill=sysTF}] coordinates {(1,0.679)(2,0.717)(3,0.819)}; \addlegendentry{Text-frame}
\addplot+[mark=square*,color=sysSU,mark options={fill=sysSU}] coordinates {(1,0.716)(2,0.758)(3,0.871)}; \addlegendentry{Seg+uniform}
\addplot+[mark=triangle*,color=sysSD,mark options={fill=sysSD}] coordinates {(1,0.720)(2,0.758)(3,0.905)}; \addlegendentry{Seg+DINO-key}
\addplot+[mark=diamond*,color=sysSF,mark options={fill=sysSF}] coordinates {(1,0.706)(2,0.730)(3,0.871)}; \addlegendentry{Seg+frame}
\addplot+[mark=o,densely dotted,line width=1.3pt,black,mark size=2pt,mark options={solid,fill=white}] coordinates {(1,0.585)(2,0.717)(3,0.890)}; \addlegendentry{No memory}
\addplot+[mark=x,densely dashed,line width=1.1pt,gray!70!black,mark size=2.2pt,mark options={solid,line width=1.3pt}] coordinates {(1,0.646)(2,0.710)(3,0.881)}; \addlegendentry{Last-seen refs}
\addplot+[mark=pentagon*,dashed,color=sysMF,mark options={fill=sysMF,solid}] coordinates {(1,0.335)(2,0.300)(3,0.238)}; \addlegendentry{MemFlow}
\addplot+[mark=triangle*,dashed,color=sysLL,mark options={fill=sysLL,solid,rotate=180}] coordinates {(1,0.207)(2,0.239)(3,0.386)}; \addlegendentry{LongLive-RAG}
\addplot+[mark=Mercedes star,dashed,color=sysIAM,mark options={solid}] coordinates {(1,0.341)(2,0.222)(3,0.186)}; \addlegendentry{IAMFlow}
\end{axis}
\end{tikzpicture}
}%
\end{minipage}\hfill
\begin{minipage}[t]{0.49\linewidth}
\centering
\resizebox{\linewidth}{!}{%
\begin{tikzpicture}[baseline=(current bounding box.north)]
\begin{axis}[
  scale only axis,
  width=0.78\linewidth,height=3.15cm,
  ybar,bar width=2.4pt,
  ymin=0.20,ymax=1.02,
  ylabel={Score},
  symbolic x coords={TF,SU,SD,SF,NM,LS,MF,LL,IAM},
  xtick=data,
  xticklabels={Text,Uni.,DINO,Frame,NoM,Last,MF,LLR,IAM},
  x tick label style={font=\tiny,rotate=45,anchor=east},
  tick label style={font=\scriptsize},
  label style={font=\scriptsize},
  title={(b) Pooled operating profiles},
  title style={font=\bfseries\scriptsize,anchor=south west,at={(0,1.04)}},
  grid=major,grid style={gray!18},
  legend cell align={left},
  legend columns=2,
  legend to name=trackb-leg-b,
  legend image code/.code={%
    \draw[#1] (0cm,-0.07cm) rectangle (0.26cm,0.07cm);%
  },
  legend style={font=\fontsize{5.6}{6.4}\selectfont,draw=none,fill=none,
    /tikz/every even column/.append style={column sep=6pt},
    row sep=1pt},
]
\addplot+[fill=blue!45,draw=blue!70!black] coordinates {(TF,.628)(SU,.689)(SD,.693)(SF,.674)(NM,.559)(LS,.639)(MF,.360)(LL,.284)(IAM,.356)}; \addlegendentry{Recall}
\addplot+[fill=orange!55,draw=orange!80!black] coordinates {(TF,.938)(SU,.915)(SD,.921)(SF,.920)(NM,.977)(LS,.964)(MF,.921)(LL,.975)(IAM,.915)}; \addlegendentry{Precision}
\addplot+[fill=green!45,draw=green!60!black] coordinates {(TF,.752)(SU,.786)(SD,.790)(SF,.778)(NM,.711)(LS,.769)(MF,.518)(LL,.440)(IAM,.513)}; \addlegendentry{F1}
\addplot+[fill=gray!55,draw=black!70] coordinates {(TF,.898)(SU,.845)(SD,.853)(SF,.858)(NM,.968)(LS,.941)(MF,.927)(LL,.984)(IAM,.920)}; \addlegendentry{Avoid-OK}
\end{axis}
\end{tikzpicture}
}%
\end{minipage}

\vspace{2pt}
\noindent
\begin{minipage}[t]{0.49\linewidth}
\centering\ref{trackb-leg-a}
\end{minipage}\hfill
\begin{minipage}[t]{0.49\linewidth}
\centering\ref{trackb-leg-b}
\end{minipage}
\caption{\textbf{Track-B generated-video scores} (same $30$-story pool as Table~\ref{tab:trackb-results}).
\textbf{(a)} Continuity recall by absence-gap bin; the dotted line is the same generator with no memory and the grey dashed line with last-seen references. Support is $N{=}701/293/210$ for near/mid/far.
\textbf{(b)} Pooled Recall, Precision, F1, and Avoid-OK.}
\label{fig:trackb-horizon}
\end{figure}
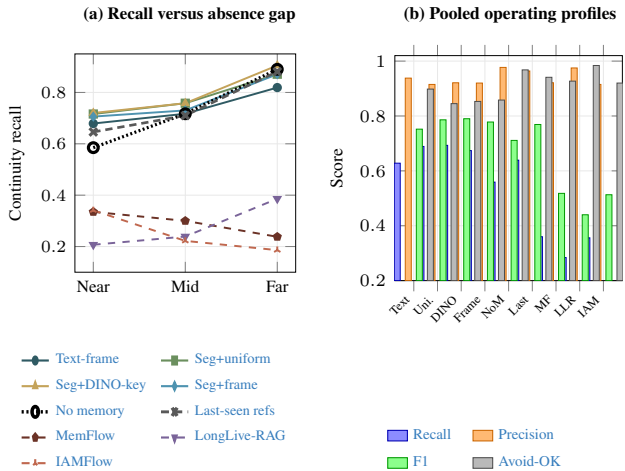

Fig.~\ref{fig:trackb-horizon} shows why a recall-versus-gap curve
misleads: every condition, including no memory, recalls more at long gaps, because a returning entity is
named in the instruction and, after a long absence, usually described (Table~\ref{tab:kind-gap}); paired differences
to no memory assess the added benefit of each complete policy.

\paragraph{Entity kind and gap.} Table~\ref{tab:kind-gap} splits recall by entity kind. The gap bins mix
kinds: returns within four segments are mostly characters named only by name, while returns after 30 or more
segments are mostly props whose names describe them and characters coming back changed with the new look in
the instruction. Memory's gain is concentrated in the first group; in the second, references add nothing
for props and hurt for changed characters.

\begin{table*}[t]
\centering
\caption{\textbf{Track-B recall by entity kind and absence gap.} No-memory recall and each condition minus no memory (paired story-bootstrap 95\% CI; gray subscripts show full CI widths (upper minus lower); bold when it excludes zero). Just seen means the entity was on screen in the previous segment. Characters are named only by name after their first appearance; 61 of the 62 character returns after 30 or more segments come back changed, with the new look in the instruction. Props often carry their look in the name. Text, Uniform, DINO-key, and Frame are the four retrieval memories.}
\label{tab:kind-gap}
\footnotesize
\setlength{\tabcolsep}{2pt}
\begin{tabular}{@{}l l r r *{5}{c}@{}}
\toprule
Kind & Gap (segments) & $n$ & \shortstack{No\\memory} & Text & Uniform & DINO-key & Frame & Last-seen \\
\midrule
Character & just seen & 2173 & 0.493 & \ci{\textbf{+0.106}}{0.07} & \ci{\textbf{+0.198}}{0.06} & \ci{\textbf{+0.201}}{0.06} & \ci{\textbf{+0.191}}{0.08} & \ci{\textbf{+0.143}}{0.07} \\[2pt]
 & 1--4 & 430 & 0.553 & \ci{\textbf{+0.144}}{0.13} & \ci{\textbf{+0.186}}{0.12} & \ci{\textbf{+0.195}}{0.12} & \ci{\textbf{+0.170}}{0.12} & \ci{+0.040}{0.12} \\[2pt]
 & 5--29 & 63 & 0.714 & \ci{-0.048}{0.24} & \ci{\textbf{+0.143}}{0.23} & \ci{+0.111}{0.25} & \ci{+0.064}{0.28} & \ci{-0.016}{0.22} \\[2pt]
 & $\geq$30 (returns changed) & 62 & 0.984 & \ci{\textbf{-0.177}}{0.21} & \ci{\textbf{-0.129}}{0.18} & \ci{-0.032}{0.13} & \ci{-0.081}{0.16} & \ci{-0.032}{0.08} \\[2pt]
\midrule
Prop & just seen & 1087 & 0.399 & \ci{\textbf{+0.070}}{0.06} & \ci{\textbf{+0.118}}{0.09} & \ci{\textbf{+0.111}}{0.08} & \ci{\textbf{+0.085}}{0.09} & \ci{\textbf{+0.049}}{0.06} \\[2pt]
 & 1--4 & 271 & 0.635 & \ci{+0.015}{0.09} & \ci{+0.044}{0.11} & \ci{+0.041}{0.11} & \ci{+0.044}{0.09} & \ci{\textbf{+0.096}}{0.11} \\[2pt]
 & 5--29 & 230 & 0.717 & \ci{+0.013}{0.10} & \ci{+0.013}{0.10} & \ci{+0.022}{0.09} & \ci{+0.000}{0.10} & \ci{-0.004}{0.09} \\[2pt]
 & $\geq$30 & 148 & 0.851 & \ci{-0.027}{0.18} & \ci{+0.027}{0.18} & \ci{+0.034}{0.19} & \ci{+0.007}{0.15} & \ci{+0.000}{0.15} \\[2pt]
\bottomrule
\end{tabular}
\end{table*}

\subsection{Identity and Historical Appearance}
\label{app:identity-results}
Full identity results are in Table~\ref{tab:identity-results}; protocols and calibration are in App.~\ref{app:identity}. Similarities condition on detected/visible pairs and may reward obsolete appearances.
\begin{table*}[t]
\centering
\caption{\textbf{Track-B identity consistency.} No-memory value and each condition minus no memory; $N$ counts eligible events, while policy-specific detected/visible subsets determine the analyzed means (bold: paired story-bootstrap 95\% CI excludes zero). Embeddings: mean cosine similarity between the previous and current appearance (DINOv3 on the localized entity; MegaLoc on the whole frame for locations), over returns detected on both sides. VLM: share judged the same individual among returns visible in both clips (lenient, Table~\ref{tab:identity-calibration}). Returns are character and prop re-appearances after at least one segment away; returns changed are characters whose look changed after 30 or more segments, for which a higher similarity means the return looks more like its old look.}
\label{tab:identity-results}
\footnotesize
\setlength{\tabcolsep}{3pt}
\begin{tabular}{@{}l r r *{5}{r}@{}}
\toprule
Stratum & \shortstack{Eligible\\$N$} & \shortstack{No\\memory} & \shortstack{Text\\$\Delta$} & \shortstack{Uniform\\$\Delta$} & \shortstack{DINO-key\\$\Delta$} & \shortstack{Frame\\$\Delta$} & \shortstack{Last-seen\\$\Delta$} \\
\midrule
\multicolumn{8}{@{}l}{\textit{DINOv3 crop similarity (trackB-embed-1.0)}}\\
All returns & 1130 & 0.35 & \textbf{+0.15} & \textbf{+0.21} & \textbf{+0.20} & \textbf{+0.21} & \textbf{+0.23} \\
Characters & 486 & 0.27 & \textbf{+0.13} & \textbf{+0.20} & \textbf{+0.19} & \textbf{+0.18} & \textbf{+0.23} \\
Props & 644 & 0.41 & \textbf{+0.16} & \textbf{+0.21} & \textbf{+0.21} & \textbf{+0.23} & \textbf{+0.22} \\
Gap 1--4 & 688 & 0.30 & \textbf{+0.15} & \textbf{+0.22} & \textbf{+0.21} & \textbf{+0.21} & \textbf{+0.25} \\
Gap 5--29 & 293 & 0.36 & \textbf{+0.16} & \textbf{+0.19} & \textbf{+0.19} & \textbf{+0.19} & \textbf{+0.17} \\
Gap $\geq$30 & 149 & 0.53 & \textbf{+0.12} & \textbf{+0.18} & \textbf{+0.17} & \textbf{+0.20} & \textbf{+0.19} \\
Returns changed, $\geq$30 & 61 & 0.19 & \textbf{+0.18} & \textbf{+0.20} & \textbf{+0.20} & \textbf{+0.15} & \textbf{+0.23} \\
\midrule
\multicolumn{8}{@{}l}{\textit{VLM pairwise judge: same individual when visible in both (trackB-identity-1.0)}}\\
All returns & 1130 & 0.84 & \textbf{+0.12} & \textbf{+0.12} & \textbf{+0.12} & \textbf{+0.12} & \textbf{+0.12} \\
Characters & 486 & 0.77 & \textbf{+0.17} & \textbf{+0.17} & \textbf{+0.15} & \textbf{+0.15} & \textbf{+0.14} \\
Props & 644 & 0.87 & \textbf{+0.11} & \textbf{+0.12} & \textbf{+0.12} & \textbf{+0.13} & \textbf{+0.12} \\
Gap 1--4 & 688 & 0.85 & \textbf{+0.10} & \textbf{+0.11} & \textbf{+0.10} & \textbf{+0.10} & \textbf{+0.10} \\
Gap 5--29 & 293 & 0.86 & \textbf{+0.11} & \textbf{+0.11} & \textbf{+0.12} & \textbf{+0.14} & \textbf{+0.11} \\
Gap $\geq$30 & 149 & 0.80 & \textbf{+0.17} & \textbf{+0.19} & \textbf{+0.18} & \textbf{+0.19} & \textbf{+0.18} \\
\midrule
\multicolumn{8}{@{}l}{\textit{ArcFace face similarity, characters}}\\
Characters & 486 & 0.13 & +0.01 & \textbf{+0.06} & \textbf{+0.05} & \textbf{+0.04} & \textbf{+0.09} \\
Returns changed, $\geq$30 & 61 & 0.06 & \textbf{+0.07} & \textbf{+0.08} & \textbf{+0.08} & \textbf{+0.08} & \textbf{+0.05} \\
\midrule
\multicolumn{8}{@{}l}{\textit{MegaLoc scene similarity, locations}}\\
Locations & 425 & 0.11 & \textbf{+0.15} & \textbf{+0.21} & \textbf{+0.18} & \textbf{+0.19} & \textbf{+0.09} \\
\bottomrule
\end{tabular}
\end{table*}

\subsection{Effects across Memory Stresses} 
\begin{figure}[t]
\centering
\includegraphics[width=\linewidth]{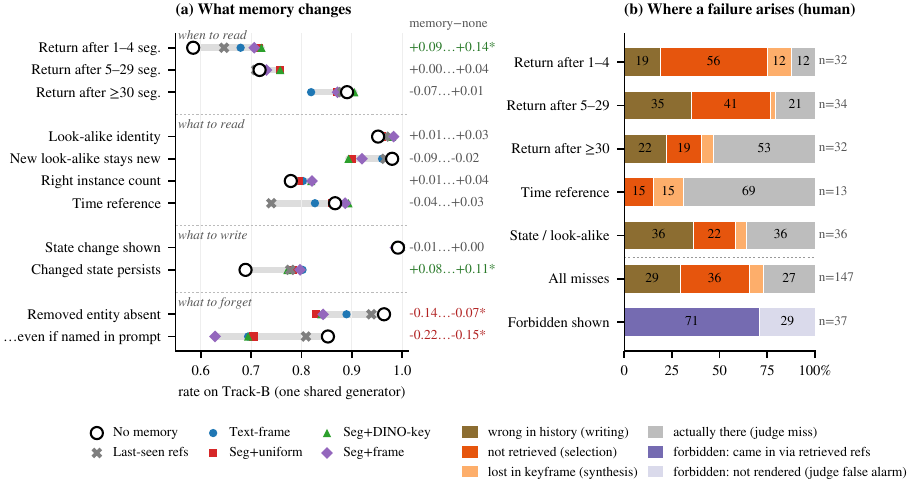}
\caption{\textbf{What memory changes, and where failures arise.}
\textbf{(a)} Track-B rate per memory stress for the shared generator with no memory (open circle) and with
each retrieval memory (filled), with last-seen references as grey crosses, grouped by the decision each stress tests (Sec.~\ref{sec:experiments}); the gap rows mix entity kinds (Sec.~\ref{sec:exp-distance}). A marker right of the circle means memory helps; left means it hurts; the
right column gives memory minus no memory (* when all four paired 95\% CIs exclude zero). Memory helps for recent returns and persistent states and hurts
where an entity must stay out.
\textbf{(b)} Human attribution of Track-B failures (App.~\ref{app:human-attribution}), by stratum: the share already
wrong in the entity's previous appearance, present there but not retrieved, retrieved but lost in the
keyframe, or actually rendered and missed by the judge. The last row splits judge-flagged forbidden renders
into those carried in by the references (purple) and judge false alarms (light purple). The two
aggregate rows are inclusion-weighted.}
\label{fig:memory-effect}
\end{figure}

Fig.~\ref{fig:memory-effect} maps every memory stress for no memory, last-seen
references, and the four retrieval memories, and repeats the human attribution by stratum.

\subsection{Per-Clip Quality Control}
\label{app:quality-control}
\bench measures memory, not visual quality; this control asks whether per-clip quality could explain
the memory differences within the shared-generator testbed. Every one of the 2,824 Track-B clips of each
condition is scored with VBench~0.1.5's own models and formulas~\citep{huang2023vbench}: imaging quality
(MUSIQ), aesthetic quality (LAION head on CLIP ViT-L/14), subject consistency (DINO), background
consistency (CLIP), dynamic degree (RAFT), and motion smoothness (AMT). Image metrics use eight fixed
frames per clip; runs are deterministic. Table~\ref{tab:quality-control} reports means with
story-clustered bootstrap 95\% CIs and the mean per-clip entity recall of the same clips. Across the four
retrieval memories the largest standardized difference in any quality metric is $0.27$, and no metric
correlates with per-clip recall above $\rho{=}0.17$ in any condition. The no-memory control has lower
recall but \emph{higher} subject and background consistency (standardized differences $0.18$--$0.45$):
these metrics reward within-clip stability, which a rollout that never recalls earlier evidence can
achieve. Released systems are omitted because they use different generators.

\begin{table*}[t]
\centering
\caption{Per-clip VBench-style quality on the shared-generator testbed (all 2,824 clips per condition).
Each cell gives the mean with its story-clustered bootstrap 95\% CI; gray subscripts show full widths (upper minus lower); Clip rec.\ is the mean
per-clip entity recall of the frozen judge on the same clips. Best in each column is \underline{underlined}.}
\label{tab:quality-control}
\small
\setlength{\tabcolsep}{2.5pt}
\begin{tabular}{@{}l ccccccc@{}}
\toprule
Condition & Imaging$\uparrow$ & Aesthetic$\uparrow$ & Subj.\ cons.$\uparrow$ & Bg.\ cons.$\uparrow$ & Dynamic$\uparrow$ & Motion sm.$\uparrow$ & Clip rec.$\uparrow$ \\
\midrule
Text-frame & \ci{0.634}{0.024} & \ci{0.584}{0.022} & \ci{0.926}{0.004} & \ci{0.930}{0.004} & \ci{0.521}{0.058} & \ci{\underline{0.9933}}{0.0005} & \ci{0.672}{0.049} \\
Seg+uniform & \ci{\underline{0.660}}{0.022} & \ci{\underline{0.596}}{0.024} & \ci{0.929}{0.005} & \ci{0.931}{0.003} & \ci{0.485}{0.071} & \ci{0.9931}{0.0006} & \ci{0.735}{0.042} \\
Seg+DINO-key & \ci{0.652}{0.023} & \ci{0.595}{0.024} & \ci{0.929}{0.004} & \ci{0.931}{0.003} & \ci{0.493}{0.073} & \ci{0.9931}{0.0005} & \ci{\underline{0.740}}{0.045} \\
Seg+frame & \ci{0.652}{0.024} & \ci{0.595}{0.023} & \ci{0.928}{0.005} & \ci{0.931}{0.003} & \ci{0.475}{0.060} & \ci{\underline{0.9933}}{0.0005} & \ci{0.731}{0.035} \\
\midrule
No memory ($K{=}0$) & \ci{0.645}{0.015} & \ci{0.573}{0.013} & \ci{\underline{0.935}}{0.004} & \ci{\underline{0.941}}{0.002} & \ci{\underline{0.549}}{0.041} & \ci{0.9932}{0.0004} & \ci{0.554}{0.068} \\
\bottomrule
\end{tabular}

\end{table*}

\subsection{Sensitivity Checks}
\label{app:sensitivity}
\label{app:robustness}
\subsubsection{Unparsable Judgments and Reference-Count Strata}
Table~\ref{tab:robustness} repeats the comparisons with no memory on segments whose judge output parsed in
both arms, since an unparsable output counts every required entity as missed and retrieval produces more of
them (91--115 segments against 55 for no memory and 79 for last-seen references). All main effects stay: gains
at 1--4 segments and for persistent states, no presence gain at 30 or more segments, and lower precision and
avoidance. Between 5 and 29 segments the three segment memories also gain significantly ($+0.05$), which
the full data leaves uncertain. The second part splits the
last-seen control by how many references it used. Both costs grow with count,
but these are unmatched segment groups: the association does not isolate a
reference-count effect. At two references, last-seen's measured costs are
about half those of retrieval on its different, three-reference sample.

\begin{table*}[t]
\centering
\caption{\textbf{Robustness of the Track-B memory effects.} Differences to no memory with paired story-bootstrap 95\% CIs; gray subscripts show full CI widths (upper minus lower); bold when the CI excludes zero. (a) Dropping every segment on which either arm's judge output failed to parse leaves all main effects in place; the three segment memories also gain significantly at 5--29 segments. (b) Last-seen effects are stratified by reference count; groups contain different segments, so this is not a count intervention.}
\label{tab:robustness}
\footnotesize
\setlength{\tabcolsep}{3pt}
\begin{tabular}{@{}l r *{6}{c}@{}}
\toprule
\multicolumn{8}{@{}l}{\textit{(a) Minus no memory, segments with a parsable judge output in both arms}}\\
Condition & dropped & Recall & Precision & Avoid-OK & 1--4 & 5--29 & $\geq$30 \\
\midrule
Text-frame & 129 & \ci{\textbf{+0.075}}{0.05} & \ci{\textbf{-0.041}}{0.02} & \ci{\textbf{-0.073}}{0.03} & \ci{\textbf{+0.105}}{0.09} & \ci{+0.015}{0.10} & \ci{-0.067}{0.13} \\[2pt]
Seg+uniform & 138 & \ci{\textbf{+0.139}}{0.05} & \ci{\textbf{-0.064}}{0.03} & \ci{\textbf{-0.128}}{0.04} & \ci{\textbf{+0.154}}{0.10} & \ci{\textbf{+0.052}}{0.09} & \ci{-0.014}{0.14} \\[2pt]
Seg+DINO-key & 143 & \ci{\textbf{+0.149}}{0.05} & \ci{\textbf{-0.057}}{0.02} & \ci{\textbf{-0.118}}{0.04} & \ci{\textbf{+0.152}}{0.10} & \ci{\textbf{+0.050}}{0.10} & \ci{+0.029}{0.11} \\[2pt]
Seg+frame & 142 & \ci{\textbf{+0.131}}{0.06} & \ci{\textbf{-0.058}}{0.02} & \ci{\textbf{-0.114}}{0.05} & \ci{\textbf{+0.143}}{0.10} & \ci{\textbf{+0.054}}{0.09} & \ci{-0.014}{0.11} \\[2pt]
Last-seen refs & 112 & \ci{\textbf{+0.087}}{0.04} & \ci{\textbf{-0.013}}{0.01} & \ci{\textbf{-0.028}}{0.02} & \ci{\textbf{+0.066}}{0.09} & \ci{-0.004}{0.09} & \ci{+0.000}{0.10} \\[2pt]
\midrule
\multicolumn{8}{@{}l}{\textit{(b) Last-seen refs minus no memory, by number of references used}}\\
References & segments & Recall & Precision & Avoid-OK & & & \\
\midrule
1 & 1803 & \ci{\textbf{+0.087}}{0.04} & \ci{-0.002}{0.02} & \ci{-0.012}{0.03} & & & \\[2pt]
2 & 842 & \ci{\textbf{+0.078}}{0.06} & \ci{\textbf{-0.027}}{0.02} & \ci{\textbf{-0.047}}{0.03} & & & \\[2pt]
3 & 149 & \ci{+0.038}{0.09} & \ci{-0.031}{0.06} & \ci{-0.052}{0.10} & & & \\[2pt]
\bottomrule
\end{tabular}
\end{table*}

\subsubsection{Independent Judge}
\label{app:second-judge}
Table~\ref{tab:second-judge} rescores the same videos with InternVL3.5-30B-A3B,
a judge from another model family, given the same roster, prompt, and clip (16 frames). It calls an entity
present far less often in every condition (no-memory recall $0.361$ against $0.559$). The two judges give the
same verdict on $88\%$ of roster entries in every memory condition and on $91\%$ without memory, where fewer
entities are on screen ($\kappa$ $0.62$--$0.67$), so their disagreement is not concentrated in any arm. Under the
second judge every memory's recall gain keeps its size ($+0.060$ to $+0.128$, every CI above zero), and the
costs of similarity retrieval keep their sign at about half the size (Avoid-OK $-0.015$ to $-0.054$, every CI
below zero; precision $-0.008$ to $-0.032$, three of four below zero), while the smaller costs of last-seen
references vanish. The per-stress effects are less stable: the gains for persistent states, the losses for
indirect references and false friends, and the last-seen loss for time references are not significant under
the second judge, so they rest on the pinned judge alone.

\begin{table}[t]
\centering
\caption{\textbf{Track-B memory effects under a second judge} (the same videos, 30 stories, same roster, prompts, and clips). Differences to no memory with paired story-bootstrap 95\% CI, gray subscripts show full CI widths (upper minus lower); bold when it excludes zero. Agreement / $\kappa$ is shown for the full roster and required-present entities.}
\label{tab:second-judge}
\scriptsize
\setlength{\tabcolsep}{1pt}
\begin{tabular}{@{}l *{3}{c} cc@{}}
\toprule
Condition & Recall & Precision & Avoid-OK & \shortstack{All\\A/$\kappa$} & \shortstack{Present\\A/$\kappa$} \\
\midrule
\multicolumn{6}{@{}l}{\itshape Qwen3-VL-32B (pinned judge)}\\
\textit{No memory} & \textit{0.559} & \textit{0.977} & \textit{0.968} & 0.908/0.64 & 0.754/0.52 \\[2pt]
Text-frame & \ci{\textbf{+0.068}}{0.04} & \ci{\textbf{-0.039}}{0.02} & \ci{\textbf{-0.070}}{0.02} & 0.882/0.64 & 0.723/0.46 \\[2pt]
Seg+uniform & \ci{\textbf{+0.130}}{0.05} & \ci{\textbf{-0.062}}{0.02} & \ci{\textbf{-0.123}}{0.05} & 0.882/0.67 & 0.727/0.45 \\[2pt]
Seg+DINO-key & \ci{\textbf{+0.133}}{0.04} & \ci{\textbf{-0.057}}{0.02} & \ci{\textbf{-0.114}}{0.03} & 0.880/0.66 & 0.713/0.43 \\[2pt]
Seg+frame & \ci{\textbf{+0.115}}{0.06} & \ci{\textbf{-0.057}}{0.02} & \ci{\textbf{-0.110}}{0.04} & 0.881/0.66 & 0.721/0.44 \\[2pt]
\textit{Last-seen refs} & \ci{\textbf{+0.080}}{0.04} & \ci{\textbf{-0.013}}{0.01} & \ci{\textbf{-0.027}}{0.02} & 0.881/0.62 & 0.700/0.43 \\[2pt]
\midrule
\multicolumn{6}{@{}l}{\itshape InternVL3.5-30B-A3B}\\
\textit{No memory} & \textit{0.361} & \textit{0.962} & \textit{0.966} &  &  \\[2pt]
Text-frame & \ci{\textbf{+0.075}}{0.03} & \ci{-0.008}{0.02} & \ci{\textbf{-0.015}}{0.02} &  &  \\[2pt]
Seg+uniform & \ci{\textbf{+0.128}}{0.05} & \ci{\textbf{-0.032}}{0.03} & \ci{\textbf{-0.054}}{0.03} &  &  \\[2pt]
Seg+DINO-key & \ci{\textbf{+0.123}}{0.05} & \ci{\textbf{-0.028}}{0.02} & \ci{\textbf{-0.050}}{0.04} &  &  \\[2pt]
Seg+frame & \ci{\textbf{+0.119}}{0.04} & \ci{\textbf{-0.024}}{0.03} & \ci{\textbf{-0.041}}{0.03} &  &  \\[2pt]
\textit{Last-seen refs} & \ci{\textbf{+0.060}}{0.04} & \ci{+0.003}{0.02} & \ci{-0.003}{0.02} &  &  \\[2pt]
\bottomrule
\end{tabular}
\end{table}

\subsubsection{Alternative Video Backends}
\label{app:second-backend} Table~\ref{tab:second-pipeline} reruns all six conditions on 8 stories
with two MiniMax-H3 realizations~\citep{minimax2026h3}. Keeping the FLUX.2 keyframe and replacing Wan2.2 with
H3 (FL2VA) reproduces every main effect with the same sign and a similar size. Removing the keyframe step, so
that H3 receives crops of the entities the instruction names from each retrieved still, reproduces the costs of
similarity retrieval (precision $-0.064$ to $-0.076$, Avoid-OK $-0.106$ to $-0.140$, every CI below zero). In
all three realizations last-seen references cost far less than retrieval. Without the keyframe, H3 partly
copies its references (near-duplicate adjacent segments), most for last-seen references, whose frames all come
from the previous shot; there its recall gain disappears, so this realization informs the retrieval costs, not
the last-seen control. Eight stories contain too few changed returns to test the stale-look effect.

\begin{table}[t]
\centering
\caption{\textbf{Track-B memory effects on two MiniMax-H3 realizations} (the same 8 stories, 767 segments per condition, seed 2026, same judge). Differences to no memory with paired story-bootstrap 95\% CI, gray subscripts show full CI widths (upper minus lower); bold when it excludes zero. Near-dup: adjacent segments whose middle frames nearly coincide (a symptom of the video model copying its references instead of following the instruction).}
\label{tab:second-pipeline}
\footnotesize
\setlength{\tabcolsep}{2pt}
\begin{tabular}{@{}l *{3}{c} c@{}}
\toprule
Condition & Recall & Precision & Avoid-OK & Near-dup \\
\midrule
\multicolumn{5}{@{}l}{\itshape FLUX.2\,$\rightarrow$\,Wan2.2 (main pipeline)}\\
\textit{No memory} (level) & \textit{0.563} & \textit{0.979} & \textit{0.969} & 0/759 \\[2pt]
Text-frame & \ci{\textbf{+0.084}}{0.05} & \ci{\textbf{-0.034}}{0.03} & \ci{\textbf{-0.067}}{0.06} & 0/759 \\[2pt]
Seg+uniform & \ci{\textbf{+0.153}}{0.06} & \ci{\textbf{-0.061}}{0.04} & \ci{\textbf{-0.130}}{0.08} & 0/759 \\[2pt]
Seg+DINO-key & \ci{\textbf{+0.150}}{0.04} & \ci{\textbf{-0.061}}{0.03} & \ci{\textbf{-0.130}}{0.07} & 0/759 \\[2pt]
Seg+frame & \ci{\textbf{+0.115}}{0.11} & \ci{\textbf{-0.067}}{0.03} & \ci{\textbf{-0.136}}{0.05} & 0/759 \\[2pt]
\textit{Last-seen refs} & \ci{\textbf{+0.071}}{0.05} & \ci{\textbf{-0.014}}{0.03} & \ci{\textbf{-0.029}}{0.04} & 0/759 \\[2pt]
\midrule
\multicolumn{5}{@{}l}{\itshape FLUX.2\,$\rightarrow$\,H3 FL2VA}\\
\textit{No memory} (level) & \textit{0.576} & \textit{0.977} & \textit{0.965} & 0/759 \\[2pt]
Text-frame & \ci{\textbf{+0.078}}{0.07} & \ci{\textbf{-0.047}}{0.03} & \ci{\textbf{-0.085}}{0.05} & 0/759 \\[2pt]
Seg+uniform & \ci{\textbf{+0.148}}{0.07} & \ci{\textbf{-0.062}}{0.03} & \ci{\textbf{-0.134}}{0.06} & 0/759 \\[2pt]
Seg+DINO-key & \ci{\textbf{+0.152}}{0.08} & \ci{\textbf{-0.046}}{0.03} & \ci{\textbf{-0.102}}{0.06} & 0/759 \\[2pt]
Seg+frame & \ci{\textbf{+0.128}}{0.09} & \ci{\textbf{-0.049}}{0.04} & \ci{\textbf{-0.100}}{0.07} & 0/759 \\[2pt]
\textit{Last-seen refs} & \ci{\textbf{+0.069}}{0.08} & \ci{-0.006}{0.03} & \ci{-0.016}{0.03} & 0/759 \\[2pt]
\midrule
\multicolumn{5}{@{}l}{\itshape H3 Ref2VA, subject crops (no keyframe)}\\
\textit{No memory} (level) & \textit{0.549} & \textit{0.988} & \textit{0.982} & 6/759 \\[2pt]
Text-frame & \ci{\textbf{+0.051}}{0.08} & \ci{\textbf{-0.064}}{0.04} & \ci{\textbf{-0.106}}{0.09} & 25/759 \\[2pt]
Seg+uniform & \ci{\textbf{+0.117}}{0.10} & \ci{\textbf{-0.076}}{0.05} & \ci{\textbf{-0.140}}{0.10} & 46/759 \\[2pt]
Seg+DINO-key & \ci{\textbf{+0.101}}{0.09} & \ci{\textbf{-0.073}}{0.06} & \ci{\textbf{-0.130}}{0.10} & 22/759 \\[2pt]
Seg+frame & \ci{\textbf{+0.095}}{0.06} & \ci{\textbf{-0.067}}{0.03} & \ci{\textbf{-0.122}}{0.06} & 53/759 \\[2pt]
\textit{Last-seen refs} & \ci{-0.037}{0.08} & \ci{\textbf{-0.016}}{0.01} & \ci{\textbf{-0.022}}{0.03} & 63/759 \\[2pt]
\bottomrule
\end{tabular}
\end{table}

\FloatBarrier

\section{Author Failure-Attribution Audit}
\label{sec:exp-human}
\label{app:human-attribution}
\subsection{Findings and Interpretation}
\label{app:audit-findings}
A missed entity may never have been rendered correctly before
(writing), not retrieved (selection), dropped by the generator (synthesis), or missed by the judge; the
testbed records every step to tell them apart. A single author inspected 224 blinded units from the four retrieval memories as an internal audit, drawn from
judge-flagged failures by failure type and gap, with judge-confirmed controls and hidden repeats
(App.~\ref{app:human-attribution}). The units, like the examples in Fig.~\ref{fig:examples}, come from an
earlier rollout of the same testbed, generator, and prompts.

Of the required-entity failures (Fig.~\ref{fig:attribution}; inclusion-weighted with 95\% CIs, raw counts 48, 38, 11,
and 50 of 147), selection accounts for 36\% (27\% to 45\%), writing for 29\% (22\% to 37\%), synthesis for
7\% (3\% to 13\%), and judge misses for 27\% (18\% to 36\%). Selection is
the part Track-A exposes; writing, a wrong generated appearance that memory faithfully stored, is an error
only the closed loop exposes. Together, 65\% of flagged misses, and 90\% of those confirmed as real, are
assigned to history or selection by the audit rule. Among 37 resolved forbidden-render units, 26 showed the entity in both
references and output (labeled context), and 11 were judge false alarms.
No unit received the generator-only label; reference co-occurrence does
not establish that the reference caused the render
(Sec.~\ref{sec:exp-trackb}).

\subsection{Audit Protocol}
\paragraph{Scope.}
A single author inspected 224 generated-output units as an internal audit; no external participants were recruited. The audit covers the four shared-generator retrieval memories in an earlier rollout than the one tabulated
in Sec.~\ref{sec:experiments}, with the same testbed, generator, and prompts. Their intermediates (references, FLUX
keyframe, Wan clip) are all recorded; released systems keep memory inside the generator's KV or latent
state, so only their final pixels could be inspected. Track-A needs no such study: no generator is in its
scored path, and its gold is human-reviewed (App.~\ref{app:bench-making}).

\begin{table*}[t]
\centering
\caption{\textbf{Where closed-loop failures arise}, by retrieval memory (\%, inclusion-weighted). Left: judge-flagged
failures of a required entity, by the stage at which it is lost: already wrong in its earlier appearance
(Hist., writing), in the history but not retrieved (Select.), in the references but lost in the keyframe
or clip (Synth.), or in fact rendered (Judge). Right: judge-flagged forbidden renders: also visible in the references
(Context), visible only in the output (Gen.), or not rendered (Judge).
These are observational stage labels, not a causal decomposition. Unresolved units excluded; pooled CIs in App.~\ref{app:audit-findings}.}
\label{tab:human-attribution}
\small
\setlength{\tabcolsep}{4pt}
\begin{tabular}{@{}l cccc c ccc c@{}}
\toprule
 & \multicolumn{5}{c}{Required entity missing or wrong} & \multicolumn{4}{c}{Forbidden entity rendered} \\
\cmidrule(lr){2-6}\cmidrule(l){7-10}
Memory & Hist. & Select. & Synth. & Judge & $n$ & Context & Gen. & Judge & $n$ \\
\midrule
Text-frame & 45 & 28 & 1 & 27 & 36 & 60 & 0 & 40 & 10 \\
Seg+uniform & 34 & 35 & 9 & 22 & 38 & 100 & 0 & 0 & 9 \\
Seg+DINO-key & 13 & 44 & 18 & 25 & 35 & 56 & 0 & 44 & 9 \\
Seg+frame & 25 & 38 & 2 & 34 & 38 & 67 & 0 & 33 & 9 \\
\midrule
Pooled & 29 & 36 & 7 & 27 & 147 & 71 & 0 & 29 & 37 \\
\bottomrule
\end{tabular}

\end{table*}

\paragraph{Frozen sample.}
From the frozen judge's per-entity verdicts we form four unit types: a required recall entity absent from
the judge's present set (\emph{miss}; stratified by absence gap 1--4, 5--29, $\ge$30, and temporal
reference), an entity present in a wrong persisted state (a state already shown earlier, so memory could
supply it) or as the wrong look-alike member (\emph{wrong}), a
destroyed or deceased entity rendered (\emph{forbidden}), and a required entity the judge found
(\emph{control}). Each retrieval memory contributes 56 units (28 miss, 10 wrong, 10 forbidden, 8 control; Table~\ref{tab:human-sample}) drawn with
seed 20260925; 10\% of units reappear later as hidden repeats, and five unscored practice items come
first. Segments without recorded references or keyframe are excluded before sampling. The manifest, media
hashes, and inclusion probabilities were written before any label was collected.

\begin{table*}[t]
\centering
\caption{\textbf{Frozen author-audit sample.} Units per retrieval memory and unit type, with the size of the
judge-flagged pool each was drawn from in parentheses; the unequal draws are why shares are
inclusion-weighted. Strata with too few candidates fall back to a neighbouring stratum (time references and
look-alike errors are pooled with mid-gap misses and state errors). The 224 units are shown with 22 hidden
repeats (246 scored items) after five unscored practice items, 251 items in all.}
\label{tab:human-sample}
\footnotesize
\setlength{\tabcolsep}{3.5pt}
\begin{tabular}{@{}l *{7}{c} r@{}}
\toprule
 & \multicolumn{4}{c}{Miss (required entity absent)} & & & & \\
\cmidrule(lr){2-5}
Memory & gap 1--4 & 5--29 & $\geq$30 & time ref. & Wrong & Forbidden & Control & Units \\
\midrule
Text-frame & 8 {\scriptsize(140)} & 8 {\scriptsize(56)} & 8 {\scriptsize(20)} & 4 {\scriptsize(8)} & 10 {\scriptsize(109)} & 10 {\scriptsize(201)} & 8 {\scriptsize(596)} & 56 \\
Seg+uniform & 8 {\scriptsize(155)} & 8 {\scriptsize(63)} & 8 {\scriptsize(12)} & 4 {\scriptsize(6)} & 10 {\scriptsize(121)} & 10 {\scriptsize(265)} & 8 {\scriptsize(582)} & 56 \\
Seg+DINO-key & 8 {\scriptsize(116)} & 10 {\scriptsize(52)} & 8 {\scriptsize(18)} & 2 {\scriptsize(2)} & 10 {\scriptsize(117)} & 10 {\scriptsize(265)} & 8 {\scriptsize(639)} & 56 \\
Seg+frame & 8 {\scriptsize(109)} & 8 {\scriptsize(42)} & 8 {\scriptsize(20)} & 4 {\scriptsize(5)} & 10 {\scriptsize(111)} & 10 {\scriptsize(274)} & 8 {\scriptsize(632)} & 56 \\
\midrule
All & 32 {\scriptsize(520)} & 34 {\scriptsize(213)} & 32 {\scriptsize(70)} & 14 {\scriptsize(21)} & 40 {\scriptsize(458)} & 40 {\scriptsize(1005)} & 32 {\scriptsize(2449)} & 224 \\
\bottomrule
\end{tabular}

\end{table*}

\paragraph{Interface and rule.}
The box below shows what each item displays, the question asked of every panel, and the deterministic rule
that turns the answers into an attribution. The entity card gives the entity's gold appearance at that
point of the story (its latest state).

\begin{tcolorbox}[vmembox,title={Human audit item}]
\textbf{Entity card.} \emph{Old Shi (character): night patrolman in his fifties, gray crew cut, deep-gray
security uniform and black leather shoes, patrol baton and key ring at the waist.} Required state or
look-alike cue, if any, in the story's original Chinese.

\textbf{Four panels, left to right.} (1) the entity's earlier appearance in the same rollout; (2) the three
references memory returned; (3) the composed keyframe; (4) the generated clip.

\textbf{Question per panel.} ``Is the correct entity visible here?'' (forbidden units: ``Is this entity
visible here?''). \textbf{Answer:} yes / no / unsure.

\textbf{Hidden from the annotator.} Memory identity, judge verdict, and unit type; required and forbidden
items differ only in the question. Items from all memories and unit types are interleaved in a fixed seeded
order; five practice items come first and hidden repeats appear in the second half.

\textbf{Attribution (deterministic, applied after labeling).} Required entity: clip shows it $\rightarrow$
judge miss; else references show it $\rightarrow$ synthesis; else earlier appearance shows it
$\rightarrow$ selection; else $\rightarrow$ writing. Forbidden entity: clip does not show it $\rightarrow$
judge false alarm; else references carry it $\rightarrow$ context; else $\rightarrow$ generator. An
``unsure'' on a deciding panel leaves the unit unresolved.
\end{tcolorbox}

\subsection{Reliability and Measurement Scope}
All 251 items were completed (median 11\,s per item). The 22 hidden repeats, which measure how
repeatable the labels are rather than agreement between annotators, received the same attribution in 20
cases; 28 of 32 judge-confirmed controls were
confirmed, and 8 of 224 units were unresolved. Per-memory shares (Table~\ref{tab:human-attribution}) have
intervals of about $\pm$20 points, so the main text uses the pooled row.
Judge misses concentrate in long-gap (17 of 32) and temporal-reference (9 of 14) failures, so Track-B
recall in these families is conservative for the retrieval memories; the study did not sample the no-memory
control, whose parse-failure rate is also lower (App.~\ref{app:bench-scoring}).

\FloatBarrier
\section{Discussion, Limitations, and Future Tests}
\label{sec:discussion}

\subsection{What the Closed-Loop Comparison Identifies}
Each policy writes its own history, so Track-B compares complete
policies. Reference-count strata are also unmatched. To isolate a
reference's effect, branch from the same story prefix and generator
state, fix the prompt, seed, and reference count, then replace one
reference with a current-state, stale-state, or distractor-filtered
alternative. Score required appearances, requested-state fidelity, and forbidden
entities separately, so any reduction in interference is read alongside
lost useful context. Use the instruction-valid state, including an older
state for an explicit flashback. These interventions remain
future work; pairing seeds alone would not identify a single-reference
effect once histories diverge.

\subsection{Design Hypotheses and Falsification}
\begin{table*}[t]
\centering
\caption{\textbf{From observations to falsifiable design hypotheses.} Proposed interventions have not been implemented or validated here. Fix generation prefix, prompt, generator state, and seed. Match count for state/filtering tests; reading gates and budget tests intentionally vary access/count and must specify removed or added references. For write policies, match the prefix before write-back and follow the resulting trajectories. A seed match alone does not fix diverged histories.}
\label{tab:directions}
\footnotesize
\setlength{\tabcolsep}{3pt}
\begin{tabular}{@{}>{\raggedright\arraybackslash}p{0.10\linewidth}
                >{\raggedright\arraybackslash}p{0.27\linewidth}
                >{\raggedright\arraybackslash}p{0.28\linewidth}
                >{\raggedright\arraybackslash}p{0.27\linewidth}@{}}
\toprule
Decision & Observed evidence & Hypothesis to test & Result counting against it \\
\midrule
When to read & Prompt stripping changes no-memory presence substantially; the stripped retrieval gain is uncertain (Table~\ref{tab:nolook}). & Gate access when history adds information beyond the prompt and baseline context, compared with always reading. & Gating loses required appearances or valid-state fidelity without reducing interference; textual detail alone is an inadequate gate. \\
Which moment & Changed returns can become more similar to their obsolete look (Table~\ref{tab:main-identity}). & Retrieve the instruction-valid entity state, allowing earlier states for flashbacks. & State-matched references do not improve requested-state correctness or damage identity continuity at the same budget. \\
How much & Last-seen costs increase across unmatched reference-count strata (Table~\ref{tab:robustness}). & Fewer references reduce interference without losing needed evidence; vary count using explicitly specified additions. & The apparent cost increase disappears under the matched comparison; additional references improve recall without an exclusion cost. \\
What to forget & Retrieval raises recall while reducing precision and avoidance (Sec.~\ref{sec:exp-trackb}). & Suppress unrequested content in useful references while retaining required identity/state cues. & Suppression fails to improve avoidance or loses useful identity/state evidence; removing more content is not automatically better. \\
What to write & The earlier single-author audit assigns 29\% of sampled flagged misses to stored wrong renderings (App.~\ref{app:human-attribution}). & Reject or repair incorrect observations before write-back. & Rejecting history discards uniquely useful evidence and worsens later recall/state fidelity compared with retaining it. \\
\bottomrule
\end{tabular}
\end{table*}

The 224-unit single-author audit uses an earlier shared-generator rollout.
Among sampled, judge-flagged required-entity misses, inclusion-weighted
shares are 29\% writing, 36\% selection, 7\% synthesis, and 27\% judge
misses (App.~\ref{app:human-attribution}). These conditional shares
are not population failure frequencies. They motivate tests of
\emph{necessity} (information absent from prompt and recent context),
\emph{temporal validity} (the state requested now, including deliberate
flashbacks), and \emph{non-interference} (unrequested content in useful
frames). Table~\ref{tab:directions} pairs each hypothesis with a result that would
count against it. A failed gate or filter would not erase the measured
prompt confound or recall/exclusion trade-off; it would show that this
particular remedy does not resolve the measurement problem.

\subsection{Limitations and Scope}
\label{app:limitations}
\paragraph{Internal validity.}
All primary Track-B conditions use one generator, prompt stream,
and scorer, but they generate distinct intermediate histories.
Thus differences estimate the effects of complete retrieval
policies on these authored stories, not the isolated causal effect
of swapping a frame at one fixed generation state. Four retrieval
policies use $K{=}3$ references; the gold-ID last-seen control
uses variable numbers of references and should not be interpreted
as a matched-budget upper bound. There is one rollout per story
and condition, so story-clustered confidence intervals do not
measure sampling across generation seeds.

\paragraph{Measurement validity.}
The pinned Qwen3-VL judge can miss entities or conflate appearance
with identity. A second, independent judge preserves the main
aggregate directions but changes absolute rates and some
fine-grained significance conclusions
(App.~\ref{app:second-judge}). Identity similarities are
conditional on visible/detected returns and can reward an old
appearance that violates the current state. Failure attribution
uses one author's stratified inspection of
an earlier rollout and does not yield a population-level
causal partition of all failure modes.

\paragraph{External validity.}
Track-B uses 30 authored, entity-labeled stories and a shared
keyframe-plus-video pipeline; the alternative video-backend check
covers eight stories. Track-A uses two real-film datasets but
does not test downstream generation on those exact segments.
Our title asks whether providing historical memory helps;
we have not established how performance varies monotonically
with storage capacity, reference count, or every type of
state-aware memory architecture. The results should not be
generalized to all genres, interactive policies, or models
without further evidence.

\paragraph{Scope of design implications.}
Table~\ref{tab:directions} proposes tests, not established remedies.
Prompt stripping does not validate a reading gate, and whole-rollout
avoidance losses do not validate content filtering. Each proposal
requires its own matched intervention and can fail while the
measurement distinctions remain useful. Broader generators, seed
variation, and interactive stories remain outside the demonstrated scope.

\end{document}